\documentclass[journal]{IEEEtran}
\usepackage{amsmath,amsfonts}
\usepackage{algorithm}
\usepackage{array}
\usepackage[caption=false,font=normalsize,labelfont=sf,textfont=sf]{subfig}
\usepackage{textcomp}
\usepackage{stfloats}
\usepackage{url}
\usepackage{verbatim}
\usepackage{graphicx}
\usepackage{cite}
\usepackage{amsmath,amssymb,amsfonts}  % For mathematical symbols and fonts
\usepackage{booktabs}  % For professional looking tables
\usepackage{ragged2e}  % For text alignment
\usepackage{float}  % To control float behavior of tables and figures
\usepackage{cite}  % For citation management
\usepackage{multirow}  % For multirow tables
\usepackage{hyperref}  % For creating hyperlinks
\usepackage{caption}  % To customize captions
\usepackage{subcaption}  % For subfigures and subtables
\usepackage{svg}  % For SVG graphics
\usepackage{algorithm}  % For algorithms
\usepackage{algpseudocode}  % For pseudocode formatting
\usepackage{enumitem}  % For better list control
\usepackage{braket}  % For quantum mechanics notation
\usepackage{array}  % For better table formatting
\usepackage{caption}
\usepackage{comment}
\usepackage{graphicx}  
\begin{document}

\title{Reward-Augmented Reinforcement Learning for Continuous Control in Precision Autonomous Parking via Policy Optimization Methods}

\author{Ahmad Suleman, Misha Urooj Khan, Zeeshan Kaleem,~\IEEEmembership{Senior Member,~IEEE}, Ali H. Alenezi, Iqra Shabbir, \\ Sinem Coleri, \IEEEmembership{Fellow,~IEEE}, Chau Yuen,~\IEEEmembership{Fellow,~IEEE} 
\thanks{The authors extend their appreciation to the Deanship of Scientific Research at Northern Border University, Arar, KSA for funding this research work through the project number NBU-FPEJ-2025-xxxx-xx.

Ahmad Suleman is with National Center for Physics (NCP), Pakistan, and Vice-chairperson Community of Research and Development (CRD) (e-mail:engineersuleman118@gmail.com)

Misha Urooj Khan is with the European Organization for Nuclear Research, CERN, Switzerland, and chairperson of CRD (e-mail: misha.urooj.khan@cern.ch)

Zeeshan Kaleem is with the Department of Computer Engineering and the Interdisciplinary Research Center for Smart Mobility and Logistics, King Fahd University of Petroleum \& Minerals (KFUPM), Dhahran 31261, Saudi Arabia (e-mail: zeeshankaleem@gmail.com)

Ali H. Alenezi is with Remote Sensing Unit, Electrical Engineering Department, Northern Border University, Arar 73213, Saudi Arabia (e-mail: ali.hamdan@nbu.edu.sa)

Iqra Shabbir is with the University of Genoa, Italy (e-mail: iqra.shabbir@edu.unige.it)

Sinem Coleri is with the Department of Electrical
and Electronics Engineering, Koc University, Istanbul (e-mail: scoleri@ku.edu.tr)

Chau Yeun is with the School of Electrical \& Electronic Engineering, Nanyang Technological University, Singapore (e-mail: chau.yuen@ntu.edu.sg)}}
\maketitle
\begin{abstract}
Autonomous parking (AP) represents a critical yet complex subset of intelligent vehicle automation, characterized by tight spatial constraints, frequent close-range obstacle interactions, and stringent safety margins. However, conventional rule-based and model-predictive methods often lack the adaptability and generalization needed to handle the nonlinear and environment-dependent complexities of AP. To address these limitations, we propose a reward-augmented learning framework for AP (RARLAP), that mitigates the inherent complexities of continuous-domain control by leveraging structured reward design to induce smooth and adaptable policy behavior, trained entirely within a high-fidelity Unity-based custom 3D simulation environment. We systematically design and assess three structured reward strategies: goal-only reward (GOR), dense proximity reward (DPR), and milestone-augmented reward (MAR), each integrated with both on-policy and off-policy optimization paradigms. Empirical evaluations demonstrate that the on-policy MAR achieves a 91\% success rate, yielding smoother trajectories and more robust behavior, while GOR and DPR fail to guide effective learning. Convergence and trajectory analyses demonstrate that the proposed framework enhances policy adaptability, accelerates training, and improves safety in continuous control. Overall, RARLAP establishes that reward augmentation effectively addresses complex autonomous parking challenges, enabling scalable and efficient policy optimization with both on- and off-policy methods. 
To support reproducibility, the code accompanying this paper is publicly available. 
%can be accessed at: \url{https://github.com/ahmadsuleman/AI-based-car-parking-using-reinforcement-learning}
\end{abstract}

\section{Introduction}
\IEEEPARstart{R}{ecent} advancements in artificial intelligence (AI)-driven control technologies have significantly accelerated progress in autonomous vehicle systems \cite{zhou}. Hence, there is rapidly increasing research interest in leveraging AI to enhance and extend the autonomous capabilities of modern vehicles   \cite{khalil2022exploiting}. Autonomous parking (AP) is a pivotal component of autonomous vehicles \cite{chen, millard2019autonomous}, characterized by high risk and minimal error tolerance, where steering control primarily administers the parking process. Unlike autonomous lane following or highway navigation, parking demands fine-grained maneuvering, obstacle avoidance, and accurate alignment within fixed spatial boundaries, such as parking slot edges. The precision involved underscores the need for the development of a control strategy in the continuous domain \cite{zhao2024automatic}. The complexity of continuous-domain control stems from the virtually infinite range of possible control actions, resulting in a highly intricate decision and exploration space. Consequently, effective solutions must handle nonlinear dynamics and high-dimensional observations \cite{shu2021driving}.

\begin{table*}[h]
\centering
\caption{Comparison of Reinforcement Learning-based Methods for Autonomous Parking.}
\label{tab:rl_parking_comparison}
\renewcommand{\arraystretch}{1.5}
\scriptsize
\begin{tabular}{@{}p{2cm}p{3cm} p{3cm} p{2.0cm} p{2.0cm} p{2cm} p{2.5cm}@{}}
\toprule
\textbf{Ref} & \textbf{Algorithm(s)} & \textbf{2D/3D Scenario} & \textbf{Action-domain} & \textbf{DRL Comp.} & \textbf{Reward Comp.} \\
\midrule
\cite{savid2023simulated} & PPO (Unity ML Agents) & 3D (Unity) & Discrete & Yes & No \\
\cite{takehara2021autonomous} & DRL (Image Seg.) & 3D (Unity) & Discrete & No & No \\
\cite{xu2024} & HRL & 2D (BEV input) & Discrete & Yes & No \\
\cite{tiong2022autonomous} & A3C-PPO  & 3D (Valet Sim) & Continuous & No & No \\
\cite{quek2021} & DQN & 3D (Unity) & Discrete & No & No \\
\cite{chan2024} & TD3 & 3D (Parking Sim) & Continuous & Yes & No \\
\cite{shaker2010} & API + Q Learning & 2D + Robot & Discrete & Yes & No \\
\cite{yang2021deep} & SAC Discrete & 3D (Urban Sim) & Discrete & No & No \\
\cite{ding2019} & KT HA Q($\lambda$) & 3D (Hill Parking) & Discrete & No & No \\
\cite{farrapo2022} & DRL (ML Agents) & 3D (Game Style Unity) & Discrete & No & No \\
\textbf{Propoed RARLAP} & On-policy, Off-policy & 3D (Unity) & Continuous & Yes & Yes \\
\bottomrule
\end{tabular}
\end{table*}

Deep reinforcement learning (DRL) has demonstrated strong potential in solving real-world problems \cite{sun2025practical}, particularly those requiring continuous-domain control. Through reward feedback, it leverages deep neural networks (DNNs) to learn high-dimensional nonlinear system dynamics. A DRL system learns through constant interactions between the agent and its environment. These interactions are evaluated using rewards, which quantify the quality of the agent’s actions and guide it toward learning an optimal policy \cite{prathiba2021hybrid}. The effectiveness of policy learning is highly dependent on the design of reward function. Therefore, informed reward modeling is crucial in ensuring policy convergence and achieving desired objectives \cite{devidze2021explicable}.

In essence, reward augmentation transforms abstract objectives into a dense and informative signal that aligns the agent's learning trajectory with the desired behavioral outcomes within high-resolution continuous domains. Embedding additional task-relevant cues into the reward signal enriches the feedback \cite{yuan2024learning}, progressively helping the agent discern between suboptimal and optimal actions, accelerating convergence, improving stability during learning, and enabling the development of more refined and human-like control policies \cite{ibrahim2024comprehensive}. DRL algorithms require a substantial amount of data for practical training; therefore, a simulation setup is crucial. Unity game engine-based 3D simulations are instrumental in generating realistic scenarios, serving as an environment for training DRL agents \cite{majumder2020deep, wang2023vision} .

Reward-Augmented Reinforcement Learning for Autonomous Parking (RARLAP) employs DRL strategies to achieve optimal vehicle steering control in the continuous domain, resulting in successful parking. Since reward is a key component in DRL, we investigate the role of reward augmentation in learning optimal parking behavior that encourages collision avoidance and improves the success rate. A sparse reward, a densely engineered reward, and a milestone-based reward function are introduced and compared in terms of their convergence and behavioral smoothness. To facilitate reproducibility and realism, we have developed an open-source custom Unity simulation-based 3D parking environment integrated with the Unity ML-Agents toolkit and an interface for deploying a custom DRL algorithm \cite{juliani2020}. The simulation features physically accurate car dynamics, sensory input for measuring distance to surrounding vehicles, and a programmable parking configuration. In this paper, the simulation environment is model-free, and a reward-augmented approach is employed to train off-policy and on-policy DRL algorithms.

\subsection{Related Studies}
AP has emerged as a core capability in intelligent transportation systems, requiring high-precision maneuvers within constrained spatial boundaries. The AP problem has been solved through rule-based \cite{yurtsever2020survey} and learning-based methods \cite{wen2022deep}. Among rule-based approaches, model-predictive control (MPC) \cite{zhao2024automatic} represents a mature and widely adopted model-based method. While MPC has proven effective in many scenarios \cite{yang}, it often faces challenges in adapting in real time to complex and dynamic environments, primarily due to nonlinear vehicle dynamics and the high dimensionality of state and action spaces \cite{kim}. Whereas learning-based methods, such as most deep learning methods, rely on perception and demonstration datasets for decision-making \cite{wen2022deep}. Consequently, DRL has gained prominence for enabling flexible and adaptive control strategies through interaction with the simulation environment \cite{sun2025practical}.

A review of the existing literature \cite{savid2023simulated}--\cite{farrapo2022} reveals that the majority of DRL-based AP systems rely on discrete control actions, which simplify the control space but severely constrain the fidelity and realism required for real-world deployment. As summarized in TABLE \ref{tab:rl_parking_comparison}, prior works relied on discretized action spaces and often lack a systematic evaluation of how reward function design impacts learning dynamics and policy performance. This oversight is particularly limiting given the well-documented sensitivity of DRL algorithms to reward structure. For instance, researchers have applied DRL to the AP problem, leveraging DRL methods such as Proximal Policy Optimization (PPO) \cite{tiong2022autonomous}, Deep Q-Networks (DQN) \cite{quek2021}, and Soft Actor-Critic (SAC) \cite{yang2021deep}, demonstrating promising results in simulated environments. However, these studies typically use discrete or low-resolution action spaces, which simplify the control problem and limit its real-world applicability. In contrast, more complex approaches, encompassing hierarchical or hybrid frameworks \cite{d2024autonomous}, utilize either path-planning modules or high-level behavior decomposition to guide parking, albeit at the cost of increased modular complexity and reduced end-to-end learning efficiency.

The reward function serves as a cornerstone of the DRL optimization framework, directly shaping the agent’s policy learning process. It not only guides behavior by encoding task objectives but also enables the agent to explore and adapt effectively within complex and dynamic environments \cite{abouelazm2024review}. Sparse reward schemes can hinder training efficiency, resulting in slow convergence or suboptimal behavior, particularly in high-dimensional continuous action spaces \cite{dayal2022reward}. Therefore, designing robust, context-aware, and well-structured reward mechanisms is crucial for enabling DRL agents to learn reliable and efficient driving behaviors in real-world autonomous driving scenarios.
To address these limitations, the proposed RARLAP framework incorporates continuous-domain control policies to enable fine-grained vehicle maneuvers, and systematically investigates the impact of reward augmentation by gradually embedding task-relevant and environment-aware feedback into the learning process.
\begin{figure*}[h]
    (a)\includegraphics[width=.90\textwidth]{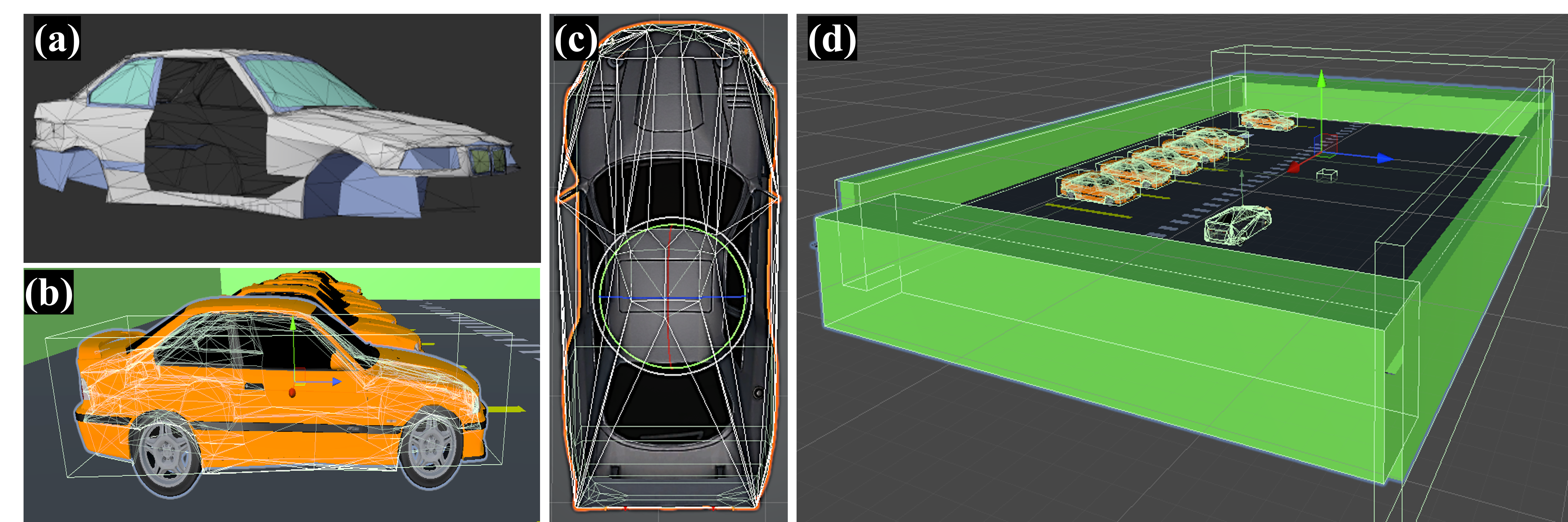}
    \caption{Unity objects with actual meshes and colliders used in environment design (a) Vehicle body mesh (b) Obstacle-vehicle’s collider and mesh (c) Agent vehicle mesh and precise collider (d) Environment colliders and meshes}
    \label{fig:env}
\end{figure*} 

\subsection{Key Paper Contributions}
Continuous-domain steering control is fundamental to enabling the effective transfer of DRL policies from simulation to reality in autonomous driving, as it captures the fine-grained actuation required for realistic maneuvers. Equally critical is the design of the reward function, which directly governs policy learning by shaping the agent’s interaction with its environment. In this context, RARLAP addresses the AP problem through a DRL-based framework in a custom 3D simulation environment. The main contributions of the paper are summarized as follows:
\begin{itemize}
    \item The existing gaps are addressed by proposing a framework that learns the control of steering angles in the continuous domain for autonomous parking without relying on expert demonstrations, rule-based planning, or modular decomposition.

    \item A systematic investigation of reward augmentation in DRL-based AP is presented, with a particular focus on its effects on policy optimization, convergence dynamics, and behavioral robustness. This research direction has received limited attention in prior work.
    
    \item A policy-focused comparison between on-policy and off-policy learning paradigms under identical rewards, simulation settings, and hyperparameters, enabling a controlled evaluation is presented. This addresses a key gap in methodological consistency across the literature and allows for a fair assessment of algorithmic performance in autonomous parking tasks.

    \item Finally, a custom Unity-based 3D autonomous parking simulator is developed and released, that incorporates realistic Ackermann steering, raycast-based distance sensing, and configurable parking scenarios. The simulator supports continuous-domain control and is designed to enable reproducible benchmarking and extensible experimentation for future DRL research. 
    %It is publicly available at \url{https://github.com/ahmadsuleman/AI-based-car-parking-using-reinforcement-learning}.

\end{itemize}
\section{Problem Formulation}\label{PF}
In this section, we formulate the DRL-based learning problem for the fixed-slot AP task using continuous-domain steering control. The approach integrates a Markov decision process (MDP) formulation for the AP, later explicitly formulating three reward strategies and on-policy and off-policy learning mechanisms.

\begin{figure}[h]
\centering
\includegraphics[width=0.49\textwidth]{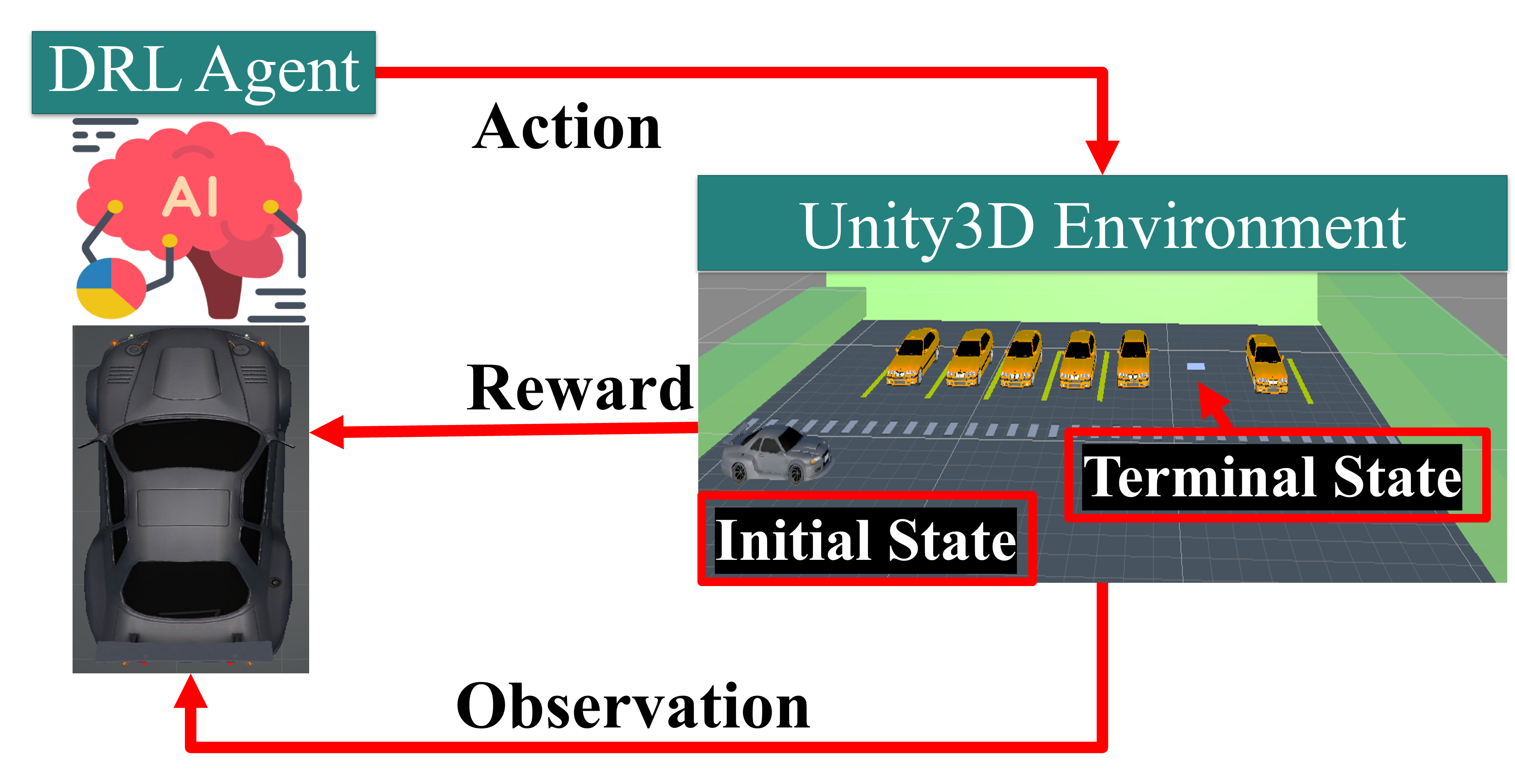}
\caption{Markov decision process environment model of autonomous parking scenario}
\label{fig:MDP_RL}
\end{figure} 

\subsection{Modeling of AP for DRL-Based Learning}
Learning the AP task through DRL algorithms requires formulating the problem as an infinite MDP, which is a sequential decision-making system that learns stochastic processes through a reward-guided strategy. An MDP is formally defined as a tuple $(\mathcal{S}, \mathcal{A}, \mathcal{P},  \mathcal{R})$, where $\mathcal{S}$ is the set of all states, $\mathcal{A}$ is the set of all actions, $\mathcal{P}(s_{t+1} | s_t, a_t)$ represents the state transition probability function, $\mathcal{R}$ is the reward function guiding the policy training. State transition probabilities are the key feature of model-based environments and are not included in model-free environments. RARLAP adopts a model-free MDP to learn steering control in the continuous domain.

Sequential decision making in DRL-based AP is shown in Fig. \ref{fig:MDP_RL}, can be described as follows: (i) at any time step \( t \), the DRL agent receives sensor information as state \( s_t \) from the 3D parking environment; (ii) a steering control action \( a_t\) in continuous domain is selected using a policy \(\pi_\theta(a_t \mid s_t) \), where \(\theta\) shows DRL agent's learning parameters; (iii) an updated state \( s_{t+1} \) of vehicle after action's influence is collected along with a reward \( r_t \) as action's feedback.

\subsubsection{State Space}
The state space $\mathcal{S}$ has been designed to include task-relevant temporal and sensory information. Each state $s_t \in \mathcal{S}$ is a real-valued vector comprising surrounding information at a given time $t$ as in (\ref{eq:state_space}).  
\begin{equation}
s_t = \left[\vec{p}^{\text{ agent}}_t, \vec{p}^{\text{ target}}_t, \vec{\mathbf{R}}_i^t , d_t\right],
\label{eq:state_space}
\end{equation}
Here, \(\vec{p}^{\text{ agent}}_t\) and \(\vec{p}^{\text{ target}}_t\) are 3D vectors containing global position coordinates of agent vehicle (AV) \((x^{\text{agent}}_t, y^{\text{agent}}_t, z^{\text{agent}}_t)\) and parking slot   \((x^{\text{target}}_t, y^{\text{target}}_t, z^{\text{target}}_t)\), respectively. \(\vec{\mathbf{R}}_i^t \in \mathbb{R}^{1 \times N}\) represents the vector of $N$ raycast range sensor readings at time \(t\), where \(i\) indexes the individual distance-ray. Therefore, \(\vec{\mathbf{R}}^t\) is computed as \([r_i^t]_{i=1}^N\), and \(d_t = \| \vec{p}^{\text{ target}}_t - \vec{p}^{\text{ agent}}_t \|_2\) is the Euclidean distance to the target.

\subsubsection{Action Space}
The action $a_t \in \mathcal{A} \subseteq [-1, 1]$ is a continuous scalar representing a normalized command to adjust the vehicle’s steering angle. At each time step \( t \), the agent selects an action \( a_t \in \mathcal{A} \), which serves as a continuous control input governing the steering angle  \( S_t^{\sphericalangle}\). Specifically, the steering angle \( S_t^{\sphericalangle}\) at time \( t \) is a function of (\ref{eq:action_space_1}), that depends upon the previous angle and the applied action as subsequent. 
\begin{equation}
S_t^{\sphericalangle} = S_{t-1}^{\sphericalangle} + \Delta S_{t-1}^{\sphericalangle}(a_t),
\label{eq:action_space_1}
\end{equation}
where \( \Delta S_{t-1}^{\sphericalangle}(a_t) \) is the steering change induced by \( a_t \).
The semantic interpretation of \( a_t \) is:
\begin{equation}
a_t =
\begin{cases}
S_t^{\sphericalangle}< S_{t-1}^{\sphericalangle} & \text{indicates clockwise steer}, \\
S_t^{\sphericalangle}= S_{t-1}^{\sphericalangle} & \text{maintains steer angle}, \\
S_t^{\sphericalangle}> S_{t-1}^{\sphericalangle} & \text{induces counter clockwise steer}.
\end{cases}
\label{eq:action_space_2}
\end{equation}
This formulation encapsulates the dynamic effect of $a_t$ on the AV’s steering in a 3D environment, enabling smooth and continuous directional adjustments.

\begin{figure*}[h]
\centering
\includegraphics[width=0.9\textwidth]{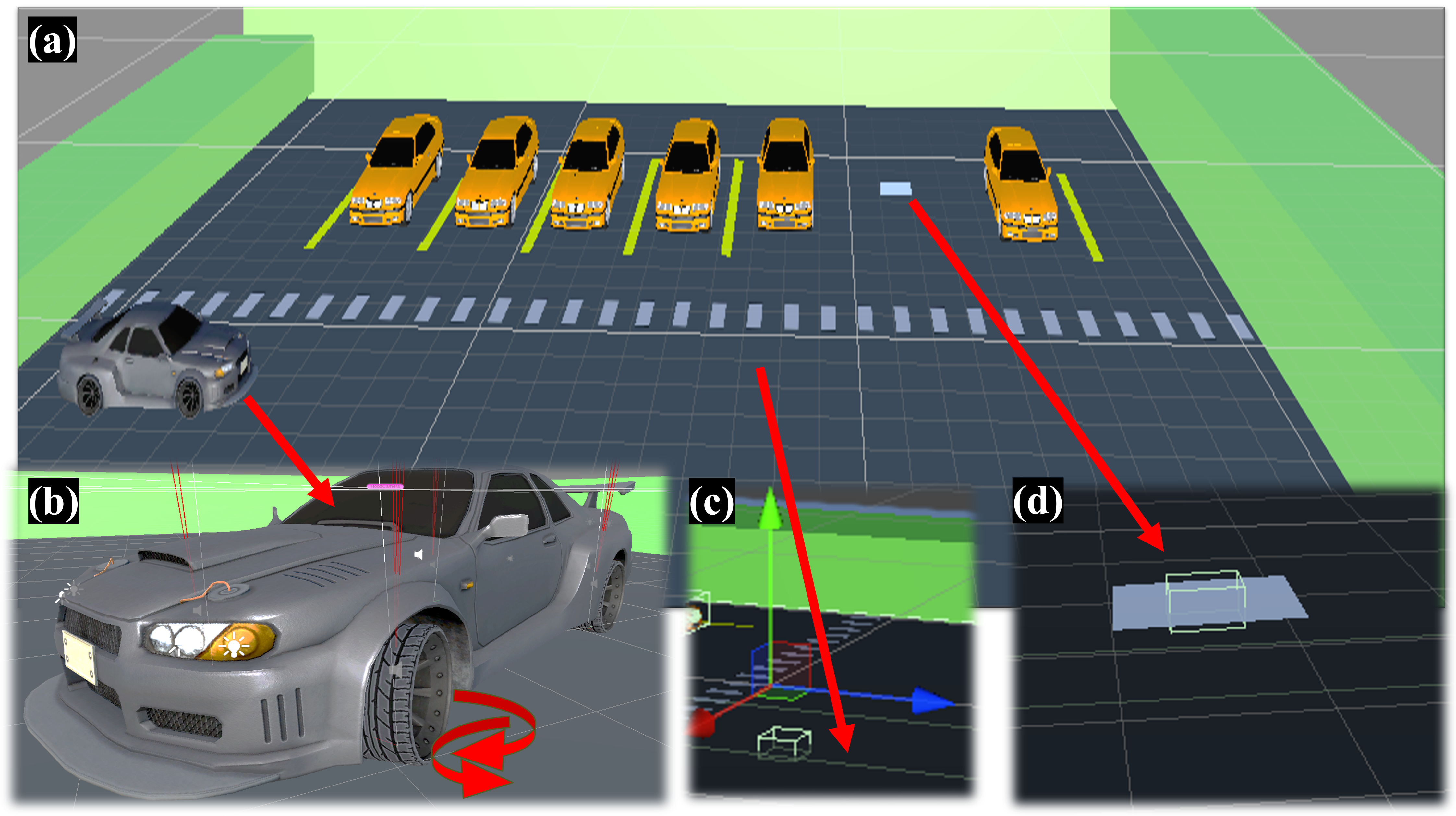}
\caption{(a) Environment layout with driving agent, occupied and empty parking slots (b) Agent’s action space control (c) Midway milestone for reward augmentation (d) Unity collider for arrival detection}
\label{EnvironmentDetails}
\end{figure*}

\begin{table*}[h]
\centering
\caption{Comparison of the Proposed Reward Strategies.}
\begin{tabular}{|l|c|c|c|c|}
\hline
\textbf{Strategy} & \textbf{Feedback Type} & \textbf{Progress Granularity} & \textbf{Safety Guidance} & \textbf{Structural Awareness} \\
\hline
GOR & Sparse         & Terminal-only        & \textcolor{red}{No}  & \textcolor{red}{No} \\
DPR & Fully Dense    & Continuous           & \textcolor{red}{No}  & \textcolor{red}{No} \\
MAR & Semi-Dense     & Post-milestone only  & \textcolor{green}{Yes} & \textcolor{green}{Yes} \\
\hline
\end{tabular}
\label{tab:all_reward_comp}
\end{table*}

\subsection{Proposed Reward Functions}
Effective reward function design is crucial to the learning performance of RL agents, particularly in goal-directed tasks that involve navigation, safety, and planning. In this section, we propose and mathematically formalize three reward strategies of increasing complexity and informativeness: 1) a sparse goal-only reward, 2) a milestone-augmented reward with hierarchical structure, and 3) a dense proximity-based reward. These strategies progressively enhance the density of reward signals and learning efficiency.
% \vspace{0.05cm}
\subsubsection{Notations}
This section introduces the primary notations used throughout the reward. 
Let the AV’s position be denoted by \( \mathbf{x}_a \in \mathbb{R}^n \), the goal position by \( \mathbf{x}_g \in \mathbb{R}^n \), and the milestone position by \( \mathbf{x}_m \in \mathbb{R}^n \). The Euclidean distance function can be computed as:
\begin{equation}
\label{eq:euclidean_distance}
D(\mathbf{x}, \mathbf{y}) = \| \mathbf{x} - \mathbf{y} \|_2.
\end{equation}
Let \( dist_{\min} \) represent the closest detected obstacle distance from the AV's sensors, and let \( \mathbb{I}[\cdot] \) denote the indicator function. We also define: minimum distance threshold \( \epsilon \) indicating a collision, radius \( M_{mil} ^{XYZ} \) within which the milestone is considered passed, and the goal region \( \mathcal{G} \subset \mathbb{R}^n \). In reinforcement learning, the effect of a reward signal on learning dynamics can be categorized as encouraging (positive) or discouraging (negative). Polarity notation is utilized to express signs of reward values. To capture this semantically, we define a polarity mapping \( \mathcal{P}(R) \), where:
\begin{equation}
\mathcal{P}(R_t) =
\begin{cases}
\overset{+}{\uparrow} & \text{if } R_t \text{ contributes positively to learning} \\
\overset{-}{\downarrow} & \text{if } R_t \text{ serves as a penalty or deterrent}.
\end{cases}
\end{equation}
\subsubsection{Sparse Goal-Only Reward (GOR)}
This reward strategy primarily provides feedback to the AP agent upon successfully reaching a parking slot or encountering a collision anywhere in the environment. The reward function is expressed as:
\begin{equation}
R_t^{GOR} =
\begin{cases}
\mathcal{F}_{\mathrm{env}}^{\,G\;\overset{+}{\uparrow}}& \text{if } \mathbf{x}_a \in \mathcal{G} \\
\mathcal{F}_{\mathrm{env}}^{\,C\;\overset{-}{\downarrow}}& \text{if } \text{dist}_{\min} \leq \epsilon \\
\mathcal{F}_{\mathrm{env}}^{\,L\;\overset{-}{\downarrow}} & \text{otherwise},
\end{cases}
\label{eq:GOR-9}
\end{equation}
where \( \mathcal{F}_{\mathrm{env}}^{G} \gg 0 \) is the successful parking reward, \( \mathcal{F}_{\mathrm{env}}^{C} \gg 0 \) is a collision penalty, and \( \mathcal{F}_{\mathrm{env}}^{L} > 0 \) is a small living penalty to discourage idle behavior and encourage exploration. The GOR strategy provides only minimal feedback on the AP environment-related aspects and will be used to assess the DRL-based AP agent's response under GOR.  

\subsubsection{Dense Proximity-Based Reward (DPR)}
The DPR strategy replaces the living penalty mechanism from the GOR feedback with a continuous, environment-aware feedback signal. Specifically, DPR provides the AP agent with a reward proportional to AV's proximity to the target parking slot, calculated as the inverse of the Euclidean distance $D$, given in (\ref{eq:euclidean_distance}), between the AV’s current position and the parking location. The DPR feedback function can be presented as:
\begin{equation}
R_t^{DPR} =
\begin{cases}
\mathcal{F}_{\mathrm{env}}^{\,G\;\overset{+}{\uparrow}} & \text{if } \mathbf{x}_a \in \mathcal{G} \\
\mathcal{F}_{\mathrm{env}}^{\,C\;\overset{-}{\downarrow}}& \text{if } \text{dist}_{\min} \leq \epsilon \\
p^\beta -p^\alpha D(\mathbf{x}_a, \mathbf{x}_g) & \text{otherwise},
\end{cases}
\label{eq:DPR-7}
\end{equation}
with \( p^\alpha > 0 \) and \( p^\beta > 0 \) as scaling parameters. This reward signal increases linearly as the AV approaches the target, providing directionally informative and dense reinforcement throughout the trajectory. Consequently, the DPR strategy enables the evaluation of the AP agent’s behavior under a continuous feedback mechanism that encodes task-relevant spatial information derived from the AP environment.

\subsubsection{Milestone-Augmented Reward (MAR)}
Building upon the GOR and DPR feedback frameworks, the MAR introduces a milestone region \( M_{mil} ^{XYZ} \), see Fig. \ref{EnvironmentDetails} (c), which serves as a spatially grounded heuristic to guide the AP agent. This region defines a feasible subset of the environment where adopting appropriate steering actions significantly increases the likelihood of achieving a successful parking maneuver. Here, $\textrm{Ind}^{M}$ is introduced as an indicator function that denotes whether the AV has entered the milestone region. It is formally defined as: 
\begin{equation}
\textrm{Ind}^{M} = \mathbb{I}[D(\mathbf{x}_a, \mathbf{x}_m) \leq M_{mil} ^{XYZ}].
\end{equation}
So, the MAR feedback function is defined as:
\begin{equation}
R_t^{MAR} =
\begin{cases}
\mathcal{F}_{\mathrm{env}}^{\,G\;\overset{+}{\uparrow}} & \text{if } \mathbf{x}_a \in \mathcal{G} \\
\mathcal{F}_{\mathrm{env}}^{\,C\;\overset{-}{\downarrow}}& \text{if } \text{dist}_{\min} \leq \epsilon \\
\zeta  - D(\mathbf{x}_a, \mathbf{x}_g) & \text{if } \textrm{Ind}^{M} = 1 \\
\mathcal{F}_{\mathrm{env}}^{\,L\;\overset{-}{\downarrow}} & \text{otherwise},
\end{cases}
\label{eq:MAR-9}
\end{equation}
Here, \( \zeta \in \mathbb{R}^+ \) denotes a shaping parameter used to ensure that the reward signal remains non-negative once the AV has passed the milestone region. The MAR strategy incorporates a geometric proximity-based feedback mechanism, structured through the milestone indicator $\textrm{Ind}^{M}$, to provide mid-trajectory guidance. The resulting reward function, \( R_t^{\text{MAR}} \), is designed to evaluate the AP agent’s learning behavior under this intermediate signal, which facilitates more effective steering control by signaling entry into a dynamically favorable region for successful parking.

The definitions of the three reward functions (TABLE~\ref{tab:all_reward_comp}) \( R^{\text{GOR}} \), \( R^{\text{DPR}} \), and \( R^{\text{MAR}} \)complete the MDP-based AP problem formulation, thereby enabling the practical evaluation of learned DRL policies under varying degrees of environment-informed feedback for an AP agent operating within a 3D parking environment.

\section{Continuous-domain Control Optimization of AP agent}
We have created an AP environment in the above section that provides us with the required information to train the AP agent for continuous-domain control. 
Here, the AP agent is tasked with learning optimal steering control strategies to safely and efficiently park the vehicle. At each discrete time step \( t \), the agent observes its current state as given in (\ref{eq:state_space}) and decides on an action as in (\ref{eq:action_space_1}) and (\ref{eq:action_space_2}). Initially, the AP agent follows a stochastic policy \( \pi_\theta(a_t \mid s_t) \), parameterized by \( \theta \), which defines a probability distribution over actions conditioned on the current state. This stochastic formulation enables structured exploration, which is particularly beneficial in continuous-domain action space as adopted in our work.

In continuous-domain AP tasks, the policy is modeled as a univariate Gaussian distribution, so the initial vehicle steering policy is defined as:
\begin{equation}
\pi_\theta(a \mid s) = \mathcal{N}(\mu_\theta(s), \sigma^2_\theta(s)),
\label{eq:random_policy}
\end{equation}
where \( \mu_\theta(s) \in \mathbb{R} \) is the mean and \( \sigma_\theta(s) \in \mathbb{R}_{>0} \) is the standard deviation of the Gaussian distribution, both predicted by a DNN conditioned on the current state \( s \).

Following the execution of the action, the environment transitions to a new state, \( s_{t+1} \). It emits a scalar reward \( r_t \), as per selected reward strategy \( R^{\text{GOR}} \) (\ref{eq:GOR-9}), \( R^{\text{DPR}} \) (\ref{eq:DPR-7}), or \( R^{\text{MAR}} \) (\ref{eq:MAR-9}). 

AP agent’s objective is to learn a steering behavior that maximizes the expected cumulative discounted reward, defined as:
\begin{equation}
J(\theta) = \mathbb{E}_{\pi_\theta} \left[ \sum_{t=0}^{\infty} \Gamma r(s_t, a_t) \right],
\label{eq:objective_function_rl}
\end{equation}
where \( \Gamma \in [0, 1) \) is the discount factor that controls the trade-off between immediate and future rewards. 

The reward formulations defined in Section \ref{PF} ensure that the AP agent receives a high positive reward upon successfully parking the vehicle without collision. Consequently, optimizing the objective \( J(\theta) \) becomes the central learning goal of the DRL-based AP agent. 
\subsection{On-Policy Optimization Mechanism (ON-POM)}
In an on-policy setting, the AP agent optimizes \( J(\theta) \) given by (\ref{eq:objective_function_rl}), based on an actor-critic mechanism, using data collected from its most recent interactions. It employs a policy gradient approach given by:
\begin{equation}
\nabla_\theta J(\theta) = \mathbb{E}_{\pi_\theta} \left[ \nabla_\theta \log \pi_\theta(a_t | s_t) \, \hat{A}_t \right],
\end{equation}
where \( \hat{A}_t \) is the advantage estimate computed using generalized advantage estimation (GAE), which reduces variance while preserving bias control.
To ensure stable updates, we optimize a clipped surrogate objective:
\begin{equation}
\mathcal{L}^{\text{CLIP}}(\theta) = \mathbb{E}_t \left[ \min\left( r_t(\theta) \hat{A}_t, \text{clip}\left( r_t(\theta), 1 - \varepsilon, 1 + \varepsilon \right) \hat{A}_t \right) \right],
\end{equation}
with \( r_t(\theta) = \frac{\pi_\theta(a_t | s_t)}{\pi_{\theta_{\text{old}}}(a_t | s_t)} \) as the importance sampling ratio, and \( \varepsilon\) controlling update conservativeness.

The associated value function $V_\Omega(s_t)$ is trained using a supervised mean-squared error objective:
\begin{equation}
\mathcal{L}^{\text{VF}}(\Omega) = \mathbb{E}_t \left[ \left( V_\Omega(s_t) - \hat{R}_t \right)^2 \right],
\label{eq:target_loss_ppo}
\end{equation}
where \( \hat{R}_t \) is the empirical return computed from the observed trajectory. This critic provides a low-variance baseline for policy updates, stabilizing the gradient estimation. To jointly optimize the actor parameters $\theta$ and critic parameters $\Omega$ in ON-POM, the following composite loss function is minimized:
\begin{equation}
\mathcal{L}_{\text{off-policy}}(\theta, \Omega) = -\mathcal{L}^{\text{CLIP}}(\theta) + c_1\,\mathcal{L}^{\text{VF}}(\Omega) - c_2\,\mathbb{E}_t \left[ \mathcal{H}\left[\pi_\theta(\cdot|s_t)\right] \right],
\label{eq:optimization_ppo}
\end{equation}
where $\mathcal{H}[\pi_\theta(\cdot|s_t)]$ denotes the policy entropy, which promotes action diversity to enhance exploration and reduce the risk of premature convergence to suboptimal policies. The scalar hyperparameters $c_1$ and $c_2$ modulate the influence of the value function loss and entropy regularization, respectively, within the total objective. Joint parameter updates are performed using stochastic gradient descent with a shared learning rate $\eta$, according to:
\begin{equation}
(\theta, \Omega) \leftarrow (\theta, \Omega) - \eta\, \nabla_{\theta,\Omega} \mathcal{L}_{\text{on-policy}}(\theta, \Omega),
\label{eq:gradient_update_ppo}
\end{equation}
This update encourages the actor to favor actions yielding higher estimated advantages under the current policy, while concurrently training the critic to improve its approximation of expected returns. Hence, training the AP agent to successfully park the vehicle. 

\subsection{Off-Policy Optimization Mechanism (OFF-POM)}
To achieve the same learning objective defined in (\ref{eq:objective_function_rl}), off-policy learning offers an alternative mechanism that improves sample efficiency and exploration capability. Unlike on-policy methods, which update the policy using freshly collected data, off-policy learning decouples data collection and policy improvement by training on transitions stored in a replay buffer. This enables the agent to learn from a broader distribution of experiences and reuse past interactions effectively.

For consistency with the parameterized on-policy formulation, we express the off-policy entropy-regularized objective as \( J(\theta) \), where \( \pi_\theta \) denotes the policy network used to approximate the policy distribution \( \pi \). The objective augments the expected cumulative reward with an entropy term that encourages stochastic exploration:
\begin{equation}
J(\theta) = \sum_t \mathbb{E}_{(s_t, a_t)} \left[ r_t + p^T \mathcal{H}(\pi_\theta(\cdot | s_t)) \right],
\end{equation}
where \( \mathcal{H}(\pi_\theta) = -\mathbb{E}[\log \pi_\theta(a_t | s_t)] \) promotes policy entropy, and \( p^T \) is a temperature coefficient dynamically tuned to match a desired target entropy. This coefficient balances exploration and exploitation to facilitate effective learning of parking strategies.

Learning in this setting relies on estimating the soft Q-function \( Q_\psi(s_t, a_t) \), which incorporates entropy into action-value estimation. The soft Q-function is optimized by minimizing the soft Bellman residual, yielding the temporal difference loss:
\begin{equation}
\mathcal{L}_Q(\psi) = \mathbb{E}_{(s_t, a_t, r_t, s_{t+1})} \left[ \left( Q_\psi(s_t, a_t) - \left( r_t + \Gamma V(s_{t+1}) \right) \right)^2 \right],
\end{equation}
where \( \Gamma \in [0, 1) \) is the discount factor and \( V(s_{t+1}) \) is the soft state value function capturing both future expected return and policy entropy:
\begin{equation}
V(s_t) = \mathbb{E}_{a_t \sim \pi_\theta} \left[ Q_\psi(s_t, a_t) - \alpha \log \pi_\theta(a_t | s_t) \right].
\end{equation}
To optimize the policy \( \pi_\theta \), we minimize the Kullback-Leibler (KL) divergence between the current policy and a Boltzmann distribution over Q-values, defined as:
\begin{equation}
\mathcal{L}_{\pi}(\theta) = \mathbb{E}_{s_t} \left[ D_{\text{KL}} \left( \pi_\theta(\cdot | s_t) \,\|\, \frac{\exp(Q_\psi(s_t, \cdot)/\alpha)}{Z(s_t)} \right) \right],
\end{equation}
Here, \( Z(s_t) \) is the partition function that ensures the distribution is normalized. The parameters of the policy network \( \theta \) and Q-function network \( \psi \) are jointly updated using stochastic gradient descent. Gradients are computed from both the policy loss \( \mathcal{L}_\pi(\theta) \) and the Q-function loss \( \mathcal{L}_Q(\psi) \), leading to the following parameter updates:
\begin{equation}
\theta \leftarrow \theta - \eta_\pi \nabla_\theta \mathcal{L}_\pi(\theta), \quad \psi \leftarrow \psi - \eta_Q \nabla_\psi \mathcal{L}_Q(\psi),
\label{eq:off-pom-update}
\end{equation}
where \( \eta_\pi \) and \( \eta_Q \) are learning rates for the policy and Q-function, respectively. These updates progressively improve the policy toward higher rewards. As a result, the DRL-based AP agent becomes increasingly proficient at generating robust and adaptive parking strategies through efficient reuse of off-policy data.

In summary, we initialized the agent with a stochastic policy as defined in (\ref{eq:random_policy}). We established both on-policy and off-policy RL paradigms for optimizing steering behavior in the AP environment. Policy updates are carried out using (\ref{eq:gradient_update_ppo}) and (\ref{eq:off-pom-update}) for on-policy and off-policy learning, respectively. Each learning paradigm is trained over $N$ iterations for all three reward formulations described in Section~III. The proposed training framework enables a rigorous, head-to-head assessment of how a steering control policy evolves toward optimality under different reward functions and learning algorithms. Our findings demonstrate that the interplay between DRL update schemes and reward design jointly governs both convergence behavior and the ultimate policy quality in the AP domain.

\section{Results and Discussion}
In this section, the design of the AP training environment, developed in Unity to closely replicate real-world parking scenarios, is presented. The precise modeling of the agent vehicle and environment components, which supports realistic interaction dynamics, is described. The parallelized training architecture, which accelerates policy learning through multiple concurrent simulations, is also introduced. The key hyperparameter settings used to ensure fair and consistent evaluation across all learning approaches are outlined, and the corresponding simulation results are presented. Finally, to encourage the reproducibility of research, the simulation code can be accessed at: \url{https://github.com/ahmadsuleman/AI-based-car-parking-using-reinforcement-learning}
\subsection{Autonomous Parking Training Configurations}
AP training environment is developed in Unity, comprising high-fidelity 3D assets that realistically represent the AP scenario \cite{unityAssetStore}. Each environment component is modeled precisely to ensure realistic interaction dynamics between the AP agent and environment during training and evaluation.

\subsubsection{Agent Vehicle (AV) Configuration} The AV is modeled using a Unity mesh comprising 1806 vertices, 1498 triangles, and 6 submeshes. These mesh-based geometries not only form the visible body but also serve as the collider surface for physical interactions. The AV's dimensions are set to realistic scale (Length $\times$ Width $\times$ Height = 4m $\times$ 2m $\times$ 2m), ensuring the simulation dynamics replicate real-world vehicular behavior.

\subsubsection{Sensor System} The AV has a lightweight perception system based on raycasting arrays. Specifically:
\begin{itemize}
    \item 8 Raycast sensors are mounted where four front-facing and four rear-facing sensors are spaced 25° apart.
    \item An additional ray is placed on each side at 65° from the outermost front ray to detect lateral proximity.
    \item Each raycast returns distance readings up to 8 meters, focusing on short-range, reactive control.
\end{itemize}
These sensor placements are intentionally minimalist in evaluating the RL agent’s capability for abrupt responses and low-latency policy adaptation.

\subsubsection{Parking Environment Components} 
The Unity environment includes basic rectangular box colliders that represent boundary walls. Small, transparent colliders are positioned at milestones and parking destinations to trigger arrival events without obstructing AV movement. Road surfaces are modeled with realistic physical parameters such as friction and gravity to preserve true-to-life ground interaction. Other vehicles in the parking environment are treated as obstacles with independent rectangular colliders, which are slightly extended beyond the original vehicle mesh to enhance safety. Fig. \ref{fig:env} and Fig. \ref{fig:MDP_RL} give a comprehensive overview of the environment and the vehicles architecture used in our AP setup. Together, these elements create a high-fidelity, sensor-equipped simulation that supports training AP agents with steering control in a continuous domain under constrained parking conditions.

\subsubsection{Parallelized Multi-Environment Training}
\begin{figure}[h]
    \includegraphics[width=.45\textwidth]{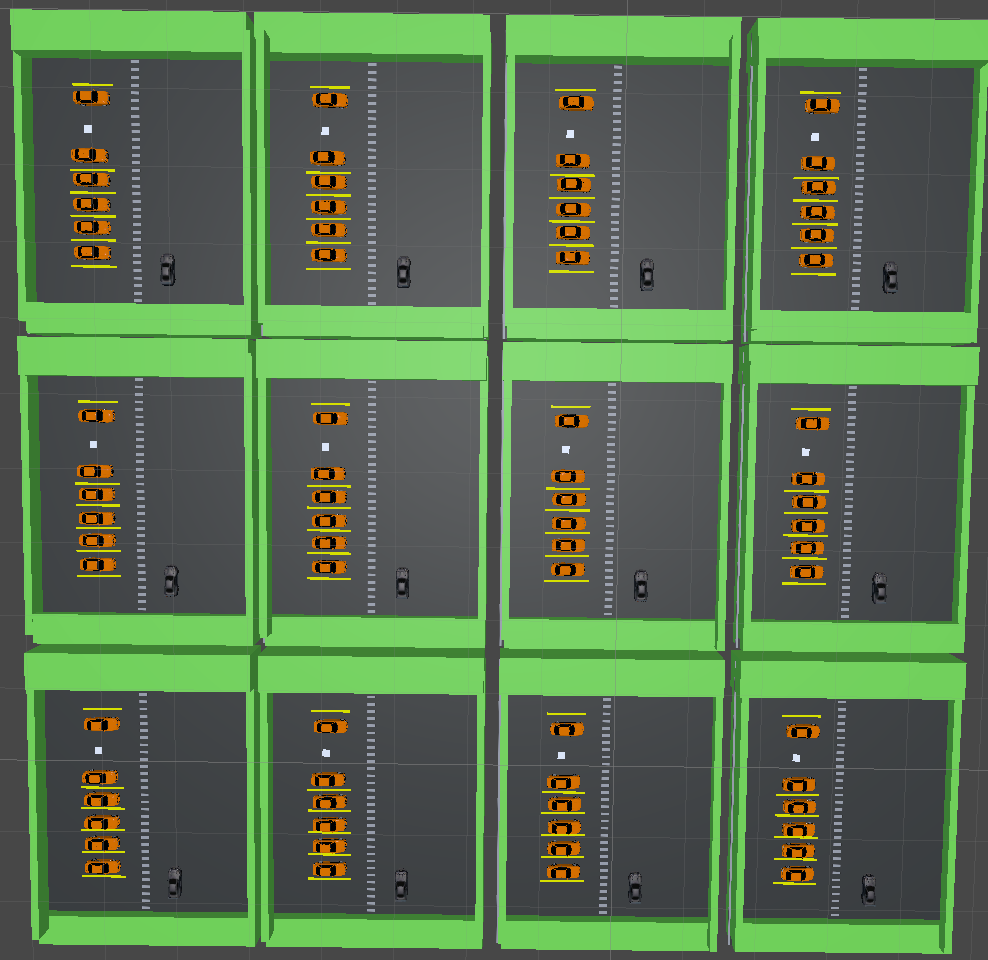}
    \caption{Illustration of parallel environment rollouts used to accelerate agent training for proposed scenarios.}
    \label{fig:parallel_env}
\end{figure}

The training of DRL-based AP agents is time-consuming and requires millions of interactions with the environment for sufficient learning. The parallel training architecture is implemented in this paper as illustrated in Fig. \ref{fig:parallel_env}, featuring 12 concurrently running instances of the Unity environment to expedite the DRL training process. Each instance simulates independent agent-environment interactions, while a centralized training module orchestrates policy updates for the selected optimization mechanism and reward function. Experiences from all environments are aggregated and used to synchronize a shared policy network. This architecture enhances data efficiency and training speed by increasing the volume and diversity of experiences, facilitating faster training and improving the generalization capabilities of the learned policy. An NVIDIA RTX 2080 GPU with 8 GB of GDDR6  memory is used for the parallel training of DRL-based AP agent. 

\subsubsection{Hyperparameters} 
As shown in TABLE~\ref{tab:hyperparams}, both ON-POM and OFF-POM share identical batch size (512) and a constant learning rate (\(\eta = 1 \times 10^{-3}\)), as well as the same discount factor ($\Gamma=0.99$), indicating equivalent temporal credit‐assignment and update granularity. OFF-POM’s inclusion of a replay buffer (\(1 \times 10^{5}\))increases sample reuse, dramatically improving data efficiency. In contrast, ON-POM’s reliance on current policy updates with clipping \( \varepsilon=0.3\) incurs additional stability constraints. 
Both ON-POM and OFF-POM employ compact 2×128 architectures, with matched parameter counts across policy $\theta$, critic $\Omega$, and value $\psi$ networks, ensuring equivalent representational capacity and inference latency. Finally, both algorithms run for 2-Million training steps. This ensures experimental consistency across optimization mechanisms, avoiding bias and allowing a fair evaluation of reward strategy impact.   

\begin{table}[h]
  \centering  
  \caption{Comparison of ON-POM and OFF-POM Hyperparameters}
  \label{tab:hyperparams}
  \begin{tabular}{@{}lll@{}}
    \toprule
    \textbf{Hyperparameter}        & \textbf{ON-POM} & \textbf{OFF-POM} \\ 
    \midrule
    Batch Size                     & 512          & 512          \\
    Replay Buffer                  & ---          & \(1 \times 10^{5}\)       \\
    Learning Rate                  & \(1 \times 10^{-3}\)       & \(1 \times 10^{-3}\)        \\
    Clipping $\epsilon$            & 0.3          & ---          \\
    Entropy Coefficient            & ---          & Auto-tuned   \\
    Discount Factor $\Gamma$       & 0.99         & 0.99         \\
    Policy Network Layers          & \(2 \times 128\) & \(2 \times 128\) \\
    Training Steps& \(2 \times 10^{6}\)    & \(2 \times 10^{6}\)    \\ 
    \bottomrule
  \end{tabular}
\end{table}

\begin{figure*}[ht]
    \centering
    (a)\includegraphics[width=.90\textwidth]{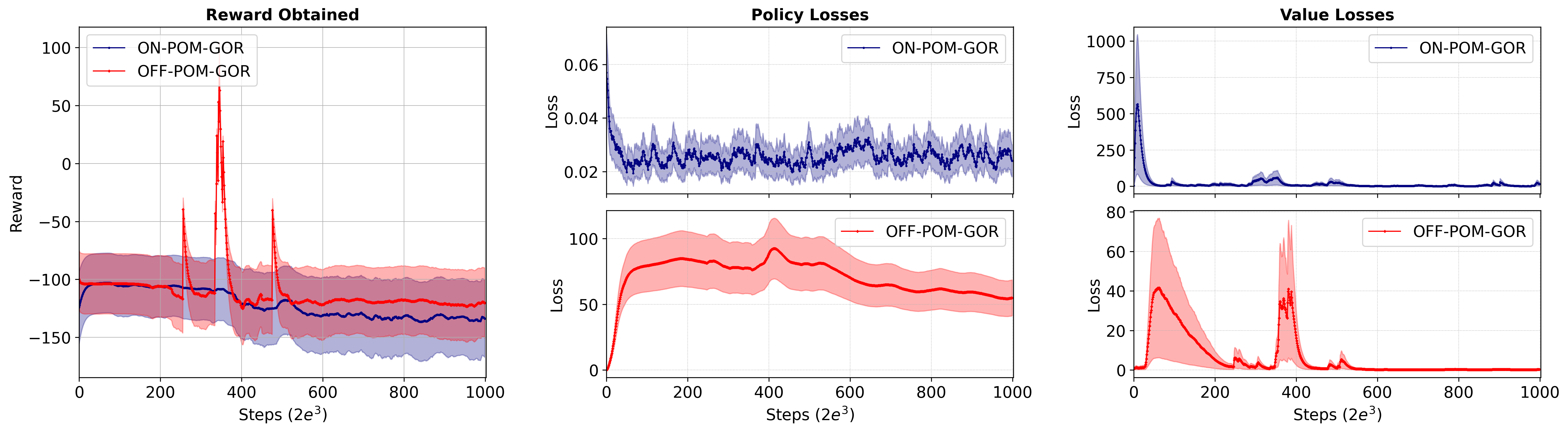}
    (b)\includegraphics[width=.90\textwidth]{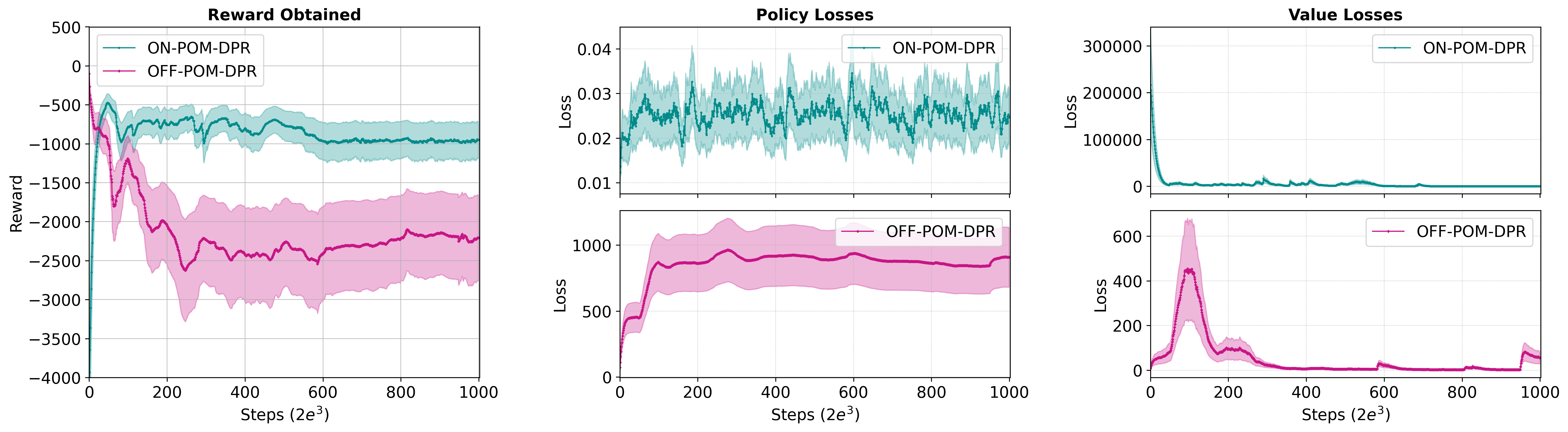}
    (c)\includegraphics[width=.90\textwidth]{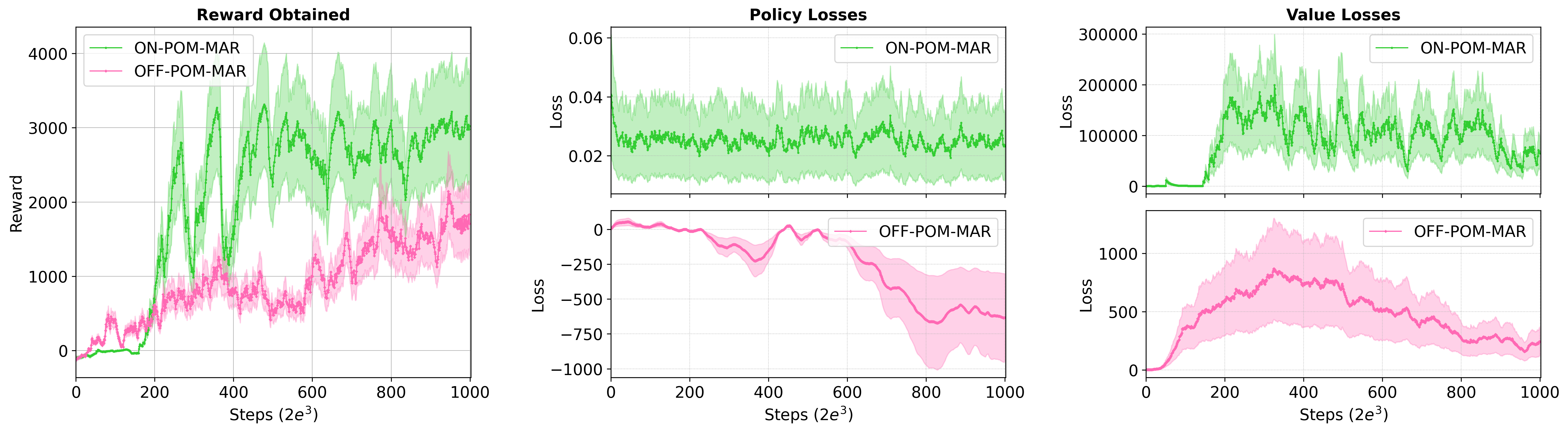}
    \caption{ON-POM and OFF-POM Training w.r.t (a) GOR (b) DPR and (c) MAR based feedback functions, policy and value losses.}
    \label{fig:benchmark}
\end{figure*} 

\begin{figure*}[ht]
    \centering
    (a)\includegraphics[width=.30\textwidth]{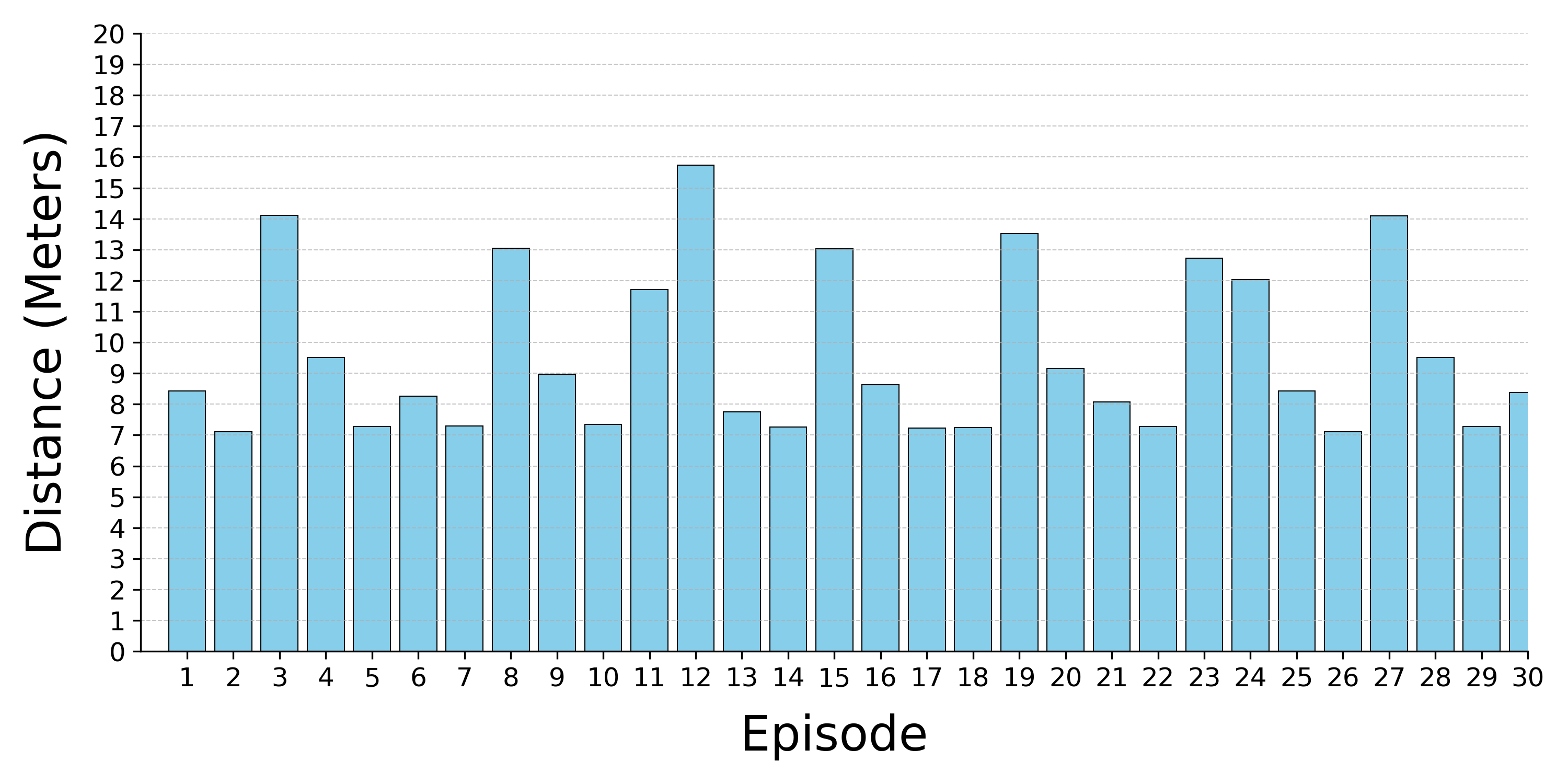}
    (b)\includegraphics[width=.30\textwidth]{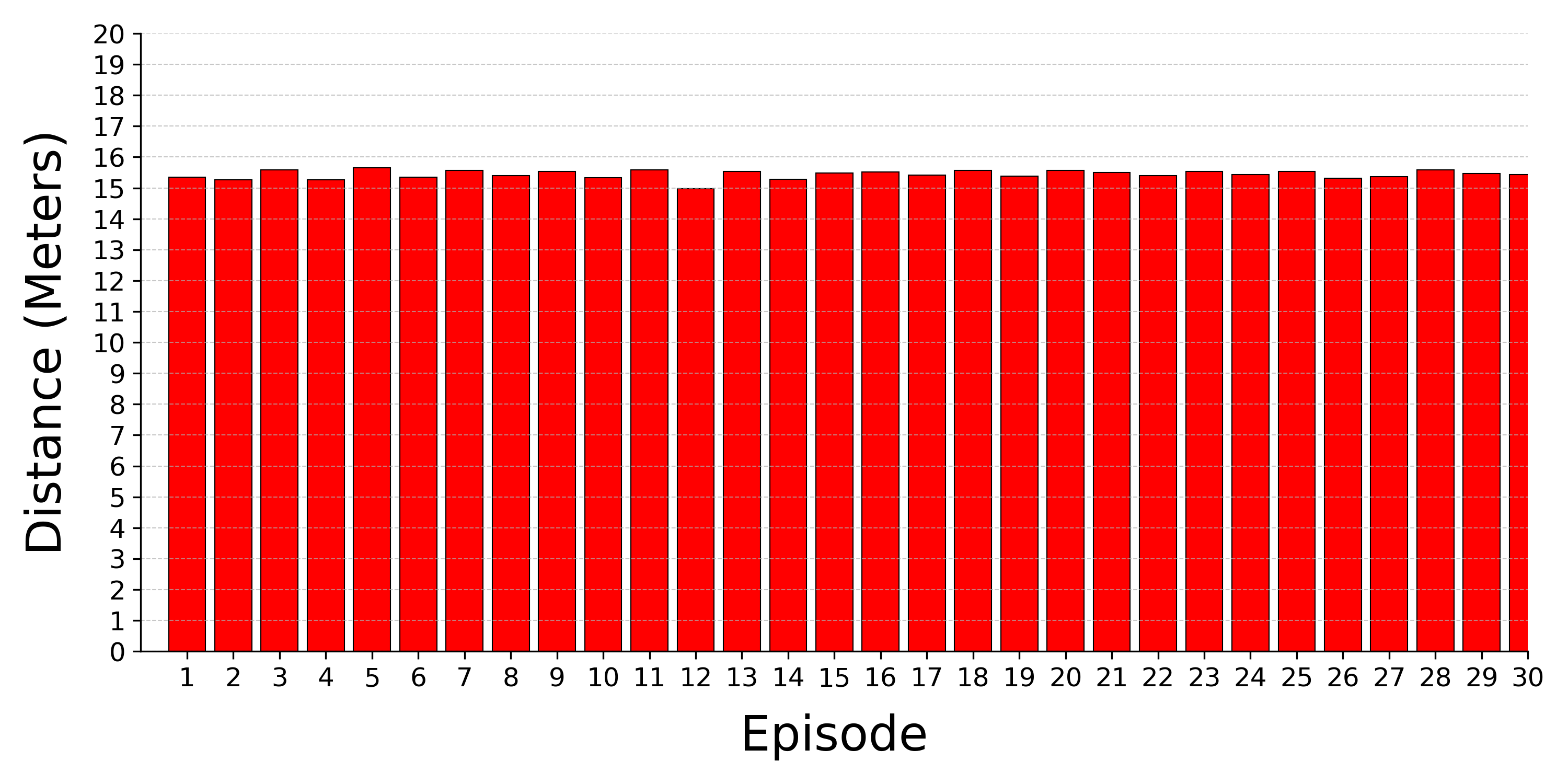}
    (c)\includegraphics[width=.30\textwidth]{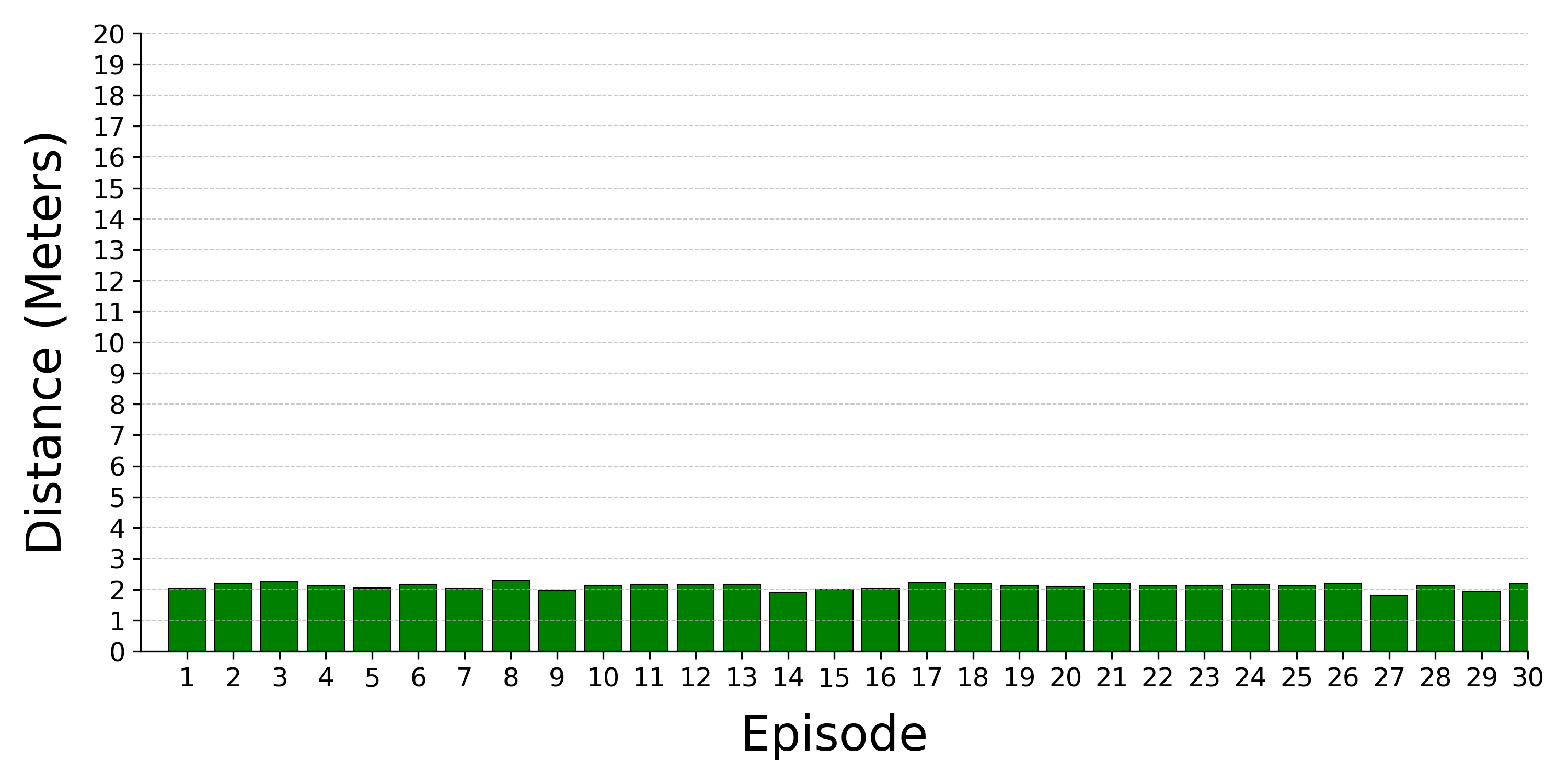}
    
    \vspace{2mm} 
    
    (d) \includegraphics[width=.25\textwidth]{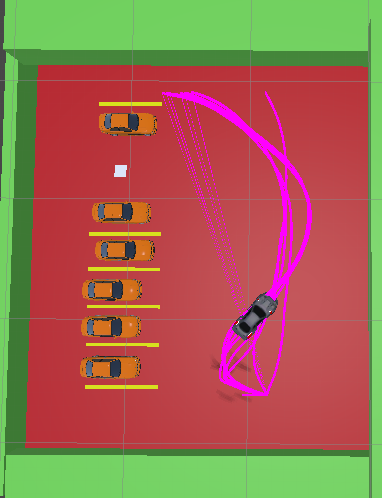}
    (e) \includegraphics[width=.25\textwidth]{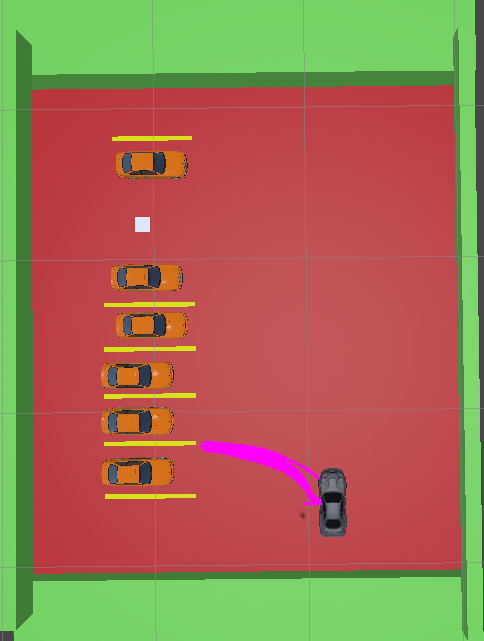}
    (f) \includegraphics[width=.25\textwidth]{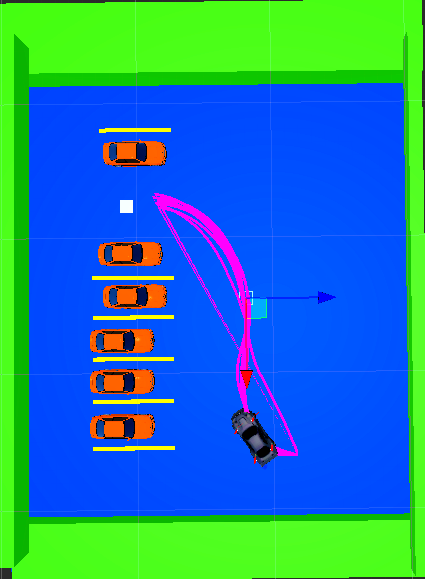}
    
    \caption{Comparison across reward strategies for best performing ON-POM AP-agent: (a–c) Euclidean distance to terminal goal under a) GOR, (b) DPR, (c) MAR ; (d–f) Trajectories adopted by ON-POM AP-agent across reward strategies (d) GOR, (e) DPR, (f) MAR.}
    \label{fig:combined_reward_eval}
\end{figure*}

\subsection{Evaluation of Reward-Based Learning Performance}

In the context of AP, where precise maneuvering, collision avoidance, and smooth control are essential, the design of the reward function plays a pivotal role in enabling DRL agents to learn reliable and efficient policies. We conducted both statistical and empirical analyses across ON-POM and OFF-POM AP-agents trained under three reward configurations, GOR, DPR, and MAR, to evaluate the impacts of reward structure on policy learning and convergence dynamics.

We begin by assessing the statistical significance of performance differences using independent two-sample Welch’s $t$-tests on 1000 reward values sampled every 2000 steps during training. This method, which does not assume equal variances, is suitable for comparing independent groups and determining whether observed differences are statistically significant.

The results strongly support the superiority of the MAR strategy. In the on-policy setting, ON-POM--MAR significantly outperforms both ON-POM--GOR ($t = 53.46$, $p \ll 0.001$, Cohen’s $d = 2.39$) and ON-POM--DPR ($t = 70.32$, $p \ll 0.001$, Cohen’s $d = 3.14$), highlighting the benefit of structured intermediate rewards in accelerating learning. Conversely, ON-POM--DPR performs significantly worse than ON-POM--GOR ($t = -116.33$, $p \ll 0.001$, Cohen’s $d = 5.20$), indicating that poorly structured dense feedback may hinder policy optimization. A similar trend is observed in the off-policy domain: OFF-POM--MAR outperforms OFF-POM--GOR ($t = 42.29$, $p \ll 0.001$, Cohen’s $d = 1.89$) and OFF-POM--DPR ($t = 112.72$, $p \ll 0.001$, Cohen’s $d = 5.04$), reinforcing the general utility of milestone-based guidance across learning paradigms.

We further investigate the influence of reward structure on convergence dynamics, as illustrated in Fig.~\ref{fig:benchmark}, by examining cumulative reward trends, policy loss, and value function stability across all configurations.

Under GOR as shown in Fig.~\ref{fig:benchmark} (a), both ON-POM and OFF-POM agents exhibit slow convergence and persistently low rewards, underscoring the challenge posed by sparse feedback in parking environments where the agent must learn from infrequent success signals. Policy loss remains relatively stable; however, OFF-POM displays intermittent spikes in value loss, indicating critic instability triggered by weak learning signals.

The DPR setting as depicted in Fig. \ref{fig:benchmark} (b) accelerates convergence but compromises final performance. ON-POM agents plateau at modest reward levels, while OFF-POM agents exhibit high reward variance and early convergence. This is accompanied by elevated policy losses and unstable value functions, indicating overestimation by the critic and erratic policy updates, dynamics that can be particularly detrimental in parking scenarios requiring stability and predictability.

In contrast, MAR shown in Fig.~\ref{fig:benchmark} (c) fosters robust learning across both paradigms. ON-POM agents rapidly achieve rewards above 3000, with consistently low policy loss and moderate value fluctuation. While OFF-POM agents also benefit from MAR, reaching rewards over 2000, they experience greater variability due to the stochastic nature of off-policy sampling. Nevertheless, both outperform their GOR and DPR counterparts, highlighting the efficacy of structured reward augmentation in guiding parking-relevant learning progression.

To further elucidate the theoretical and practical strengths of MAR, we analyze behavioral and control-theoretic implications as plotted in Fig. \ref{fig:combined_reward_eval}. From a control perspective, MAR serves as a potential field, encouraging smooth and dynamically feasible trajectories, as manifested in the consistent, goal-directed paths shown in Fig. \ref{fig:combined_reward_eval} (f). In contrast, GOR and DPR result in erratic or oscillatory behaviors shown in Fig. \ref{fig:combined_reward_eval} (d) and Fig. \ref{fig:combined_reward_eval} (e), which are unsuitable for confined, obstacle-rich environments typical of AP.

From an information-theoretic standpoint, MAR reduces policy entropy through structured intermediate signals, resulting in lower-variance learning and tighter outcome clustering as shown in Fig. \ref{fig:combined_reward_eval} (c). GOR, by contrast, produces high variance due to sparse updates plotted in Fig. \ref{fig:combined_reward_eval} (a), while DPR converges consistently but ineffectively as in Fig. \ref{fig:combined_reward_eval} (b). These dynamics align with the critical requirements of AP systems, where precise control, low-risk exploration, and policy reliability are essential for real-world deployment.

In summary, reward structure critically shapes both policy outcomes and convergence behavior. Among the evaluated strategies, milestone-based reward augmentation consistently enables faster learning, greater policy stability, and more reliable performance in safety-critical tasks such as AP. On-policy agents, in particular, are shown to exploit this structure more effectively, benefiting from the synergy between reward augmentation and synchronous update mechanisms. These findings highlight the crucial role of reward design in synchronizing learning dynamics with the operational requirements of autonomous vehicle maneuvering and positioning MAR as a robust and scalable approach for real-world DRL-based parking solutions.

\begin{figure}[h]
\centering
\includegraphics[width=0.5\textwidth]{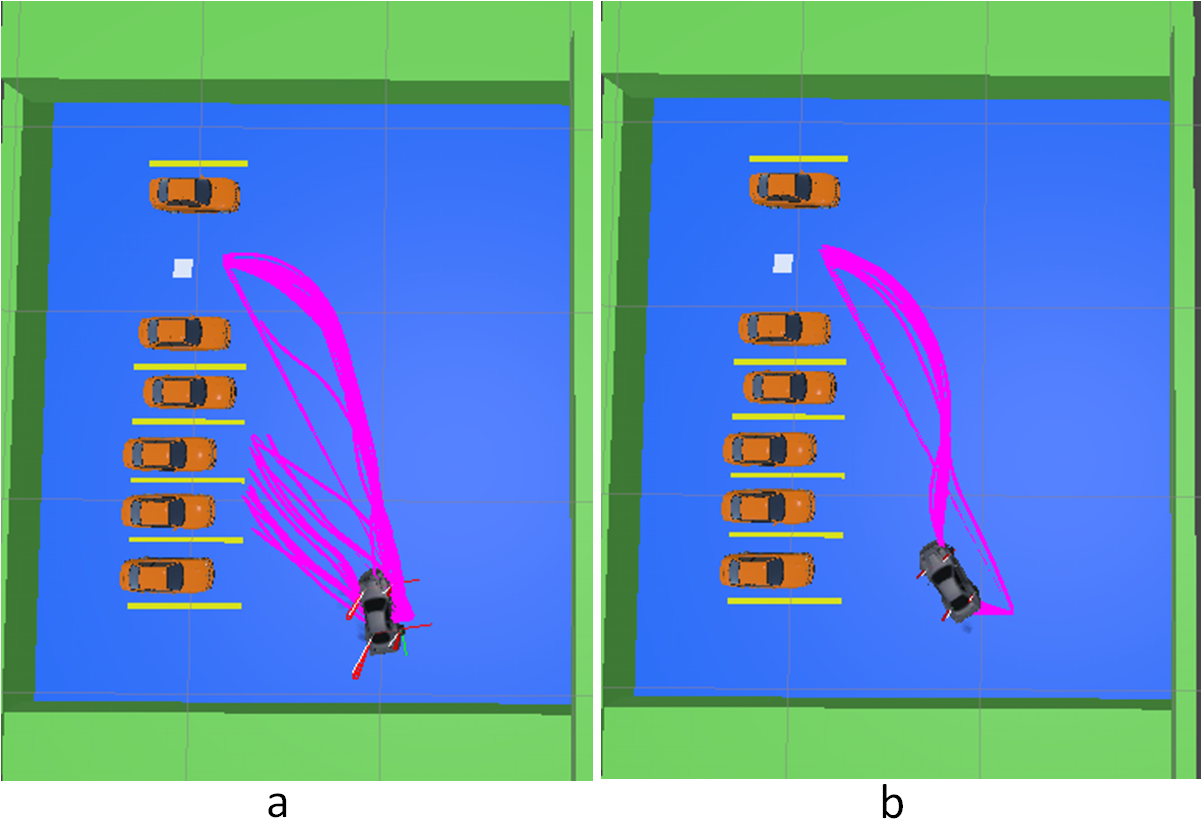}
\caption{Policy smoothness comparison through learned trajectories by (a) OFF-POM and (b) ON-POM under the MAR reward strategy.}
\label{fig:success_rate_compare}
\end{figure}

\begin{table*}[ht]
\centering
\caption{Performance comparison between ON-POM and OFF-POM agents under the MAR strategy.}
\label{tab:deployment_metrics}
\begin{tabular}{lccccc}
\toprule
\textbf{Agent–Reward} & \textbf{Success Rate (\%)} & \textbf{Collision Rate (\%)} & \textbf{Avg. Final Distance (m)} & \textbf{Training Time (min.)} & \textbf{Avg. Steps/Episode} \\
\midrule
\textbf{ON-POM (MAR)} & \textbf{91} & \textbf{9} & \textbf{2.191} & \textbf{26.83} & \textbf{201} \\
\textbf{OFF-POM (MAR)} & 50 & 50 & 7.287 & 268.74 & 143 \\
\bottomrule
\end{tabular}
\vspace{-0.6cm}
\end{table*}

\subsection{Learning Mechanism based Comparative Analysis of AP-agents}
In the context of AP under the MAR reward strategy, a comparative analysis of ON-POM and OFF-POM-based training reveals substantial differences in policy quality, deployment effectiveness, and training efficiency. As depicted in Fig \ref{fig:success_rate_compare}, the ON-POM AP-agent produces smoother and more deterministic trajectories with minimal deviation, directly correlating with its higher success rate (91\%) and lower collision rate (9\%), as shown in TABLE~\ref{tab:deployment_metrics}.  In contrast, the OFF-POM AP-agent exhibits erratic and dispersed paths, indicative of higher policy variance and less stable convergence, resulting in a lower success rate (50\%) and a significantly higher collision rate (50\%).

The ON-POM AP-agent also achieves a lower average final distance to the parking goal (2.191 m vs. 7.287 m), demonstrating superior control precision. Although it completes episodes with more steps on average (201 vs. 143), viewed in conjunction with its consistently high success rate, this behavior indicates that the policy executes stable, purposeful steering rather than engaging in inefficient or prolonged exploration. Conversely, the lower step count in OFF-POM, paired with erratic trajectories and poor outcomes, reflects an unstable or prematurely terminating policy.
Training time analysis further highlights a key trade-off: OFF-POM requires significantly longer training durations (268.74 minutes vs. 26.83 minutes), mainly due to the added per-step computational overhead introduced by its use of a replay buffer, soft target updates, and delayed policy optimization after initial random interactions.

In summary, the experimental results clearly establish the superiority of the ON-POM AP-agent when trained under the MAR feedback strategy, delivering consistently higher policy stability, improved deployment robustness, and markedly better sample and time efficiency compared to its off-policy counterpart. While the OFF-POM agent achieved competitive performance, its extended training time (268.74 minutes vs. 26.83 minutes) reflects the inherent overhead introduced by experience replay, soft target updates, and delayed policy optimization, particularly during early training phases. Given the domain-specific complexity of our custom environment, direct performance comparisons against standardized benchmarks were not applicable. However, to ensure methodological rigor, we conducted a comprehensive intra-environment evaluation across three reward strategies that progressively incorporate environment-aware information. These findings empirically validate our central hypothesis that integrating structured, environment-aware signals into the reward function is pivotal for improving learning efficiency and policy robustness, particularly in precision-sensitive autonomous driving tasks.

\section{Conclusion}
This work introduced RARLAP, a reward-augmented DRL framework tailored for precision AP under continuous-domain control diversity. By systematically evaluating three reward strategies GOR, DPR, and MAR, we demonstrate that structured intermediate rewards are critical to effective policy learning. Among the tested configurations, ON-POM combined with MAR consistently achieved the highest success rate, smoothest trajectories, and lowest collision rates, confirming the synergy between on-policy updates and structured augmentations. Conversely, both GOR and DPR failed to support reliable learning, underscoring the inadequacy of sparse or poorly structured signals in safety-critical tasks. Our findings establish that reward design is not merely a tuning parameter but a fundamental driver of learning stability and behavioral quality. Looking ahead, the integration of curriculum learning, adaptive reward scaling, and real-world sim-to-real transfer will further bridge the gap between simulated success and field-ready deployment for AP systems.

\begin{IEEEbiography}[{\includegraphics[width=1in,height=1.25in,clip,keepaspectratio]{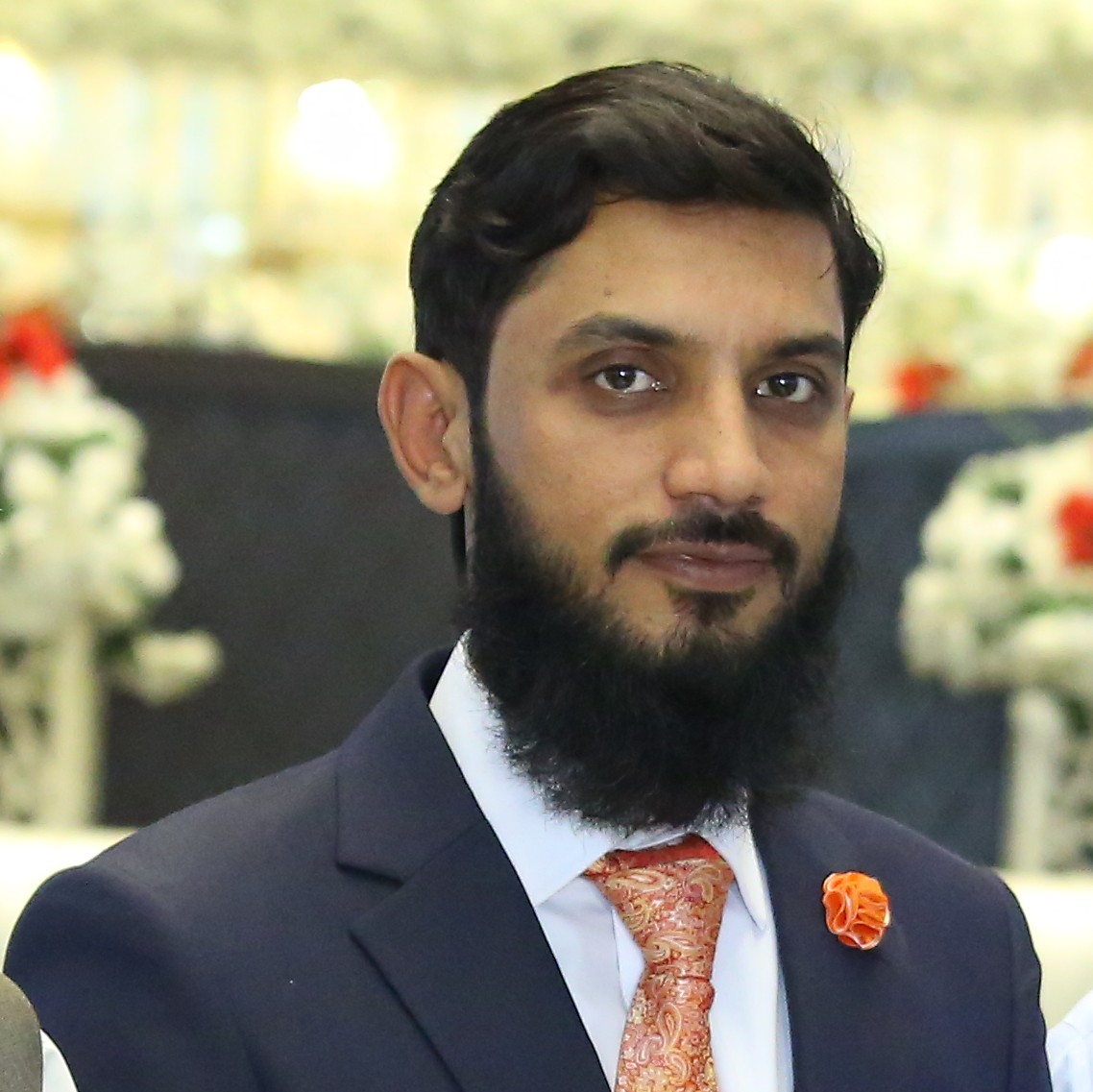}}]{Ahmad Suleman} is currently serving in the National Centre for Physics (NCP)  as Assistant Manager and as vice chairperson in the community of research and development (CRD). He holds an M.Phil. in Microelectronic Engineering and Semiconductor Physics from the University of the Punjab and a B.Sc. in Electronic Engineering from the Islamia University of Bahawalpur. He actively serves as a reviewer for IEEE Transactions on Vehicular Technologies and Springer journals. He is also an experienced AI freelancer and open-source contributor, with 10 peer-reviewed publications in IEEE and AI-focused journals. His research interests span deep reinforcement learning, autonomous robotic systems, embedded AI, computer vision, and sim2real transfer for intelligent navigation. He has developed and deployed advanced AI models across quadrupeds, hexapods, and ROS-based platforms, and has contributed to national training programs for industrial robotics. 
\end{IEEEbiography}

\begin{IEEEbiography}[{\includegraphics[width=1in,height=1.25in,clip,keepaspectratio]{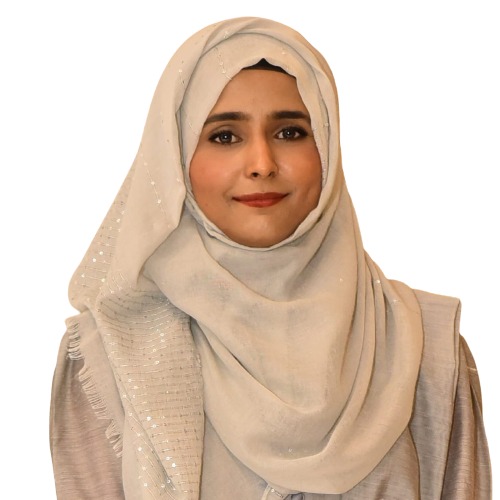}}]{Misha Urooj Khan} is a software engineer at the European Organization for Nuclear Research (CERN), Switzerland, and also the chairperson of the community of research and development (CRD). She holds a B.S. and M.S. degree in Electronics Engineering from the University of Engineering and Technology (UET), Taxila. She has published 40 international articles in the fields of quantum machine learning, quantum error mitigation, post-quantum cryptography, predictive maintenance, machine/deep learning, large language models, embedded systems, audio processing, image processing, computer vision, and biomedical signal processing. She has the honor of being the reviewer of multiple prestigious QI-ranked journals \textit{Elsevier - Computer \& Electrical Engineering, Expert System with Applications, and Engineering Applications of Artificial Intelligence}and \textit{Nature Scientific Reports.}
\end{IEEEbiography}

\begin{IEEEbiography}[{\includegraphics[width=1in,height=1.25in,clip,keepaspectratio]{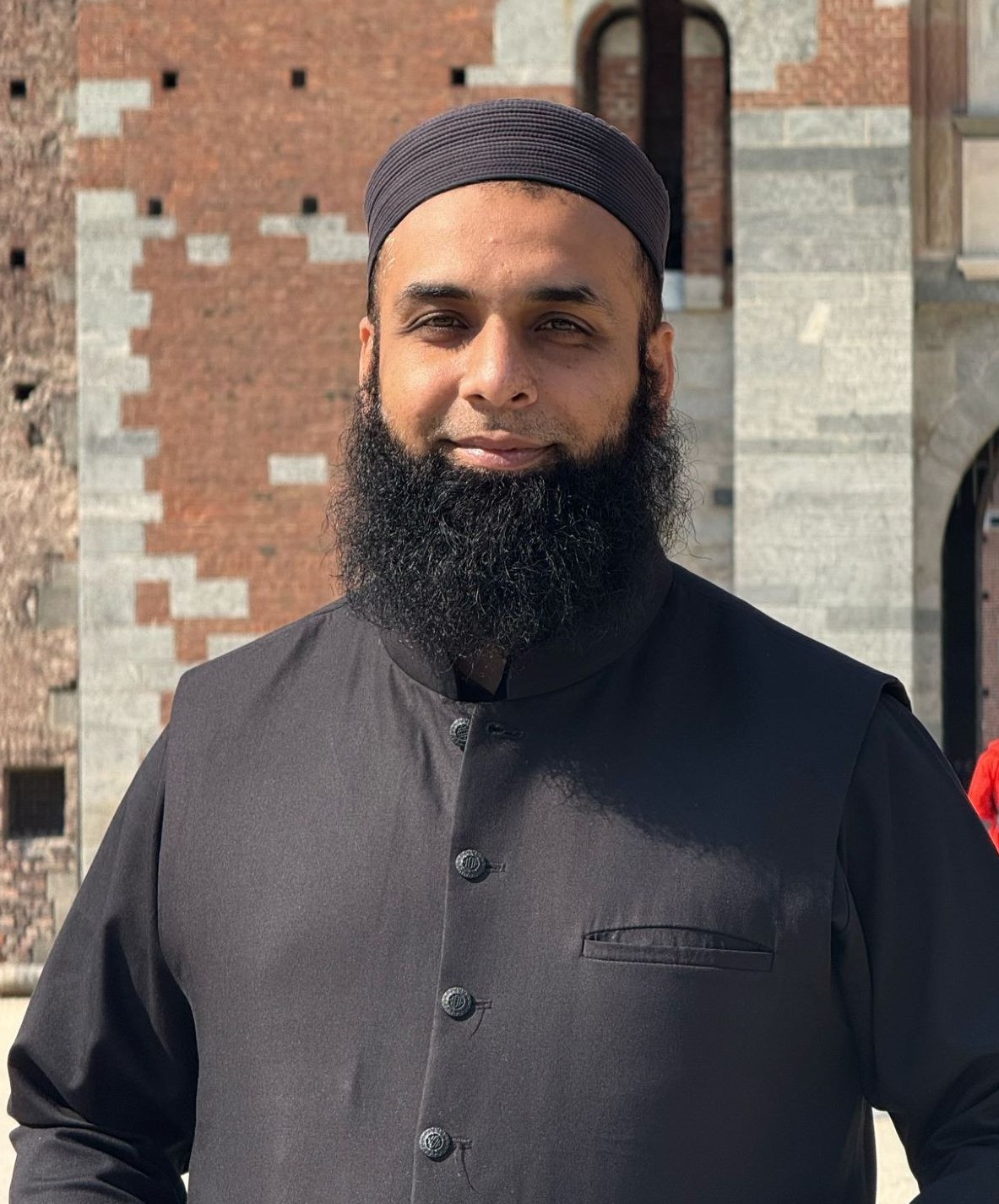}}]{Zeeshan Kaleem} (Senior Member, IEEE) is serving as an Assistant Professor in the Computer Engineering Department, King Fahd University of Petroleum and Minerals (KFUPM), Saudi Arabia. Prior to joining KFUPM he served for 8 Years at COMSATS University Islamabad. He received his BS from UET Peshawar in 2007. He received MS and Ph.D. in Electronics Engineering from Hanyang University, and Inha University, South Korea in 2010 and 2016, respectively. Dr. Zeeshan consecutively received the National Research Productivity Award (RPA) awards from the Pakistan Council of Science and Technology (PSCT) in 2017 and 2018. We won the Runner-up Award in the National Hackathon 23 competition for Project to develop Drone Detection system.  He won the Higher Education Commission (HEC) Best Innovator Award in 2017, with a single award from all over Pakistan. He received the 2021 Top Reviewer Recognition Award for IEEE Transactions on Vehicular Technology. He has published 100+ technical journal papers, including 21 as 1st author papers, books, book chapters, and conference papers in reputable journals/venues, and holds 21 US and Korean patents. He has also received research grants of around 70k US\$. He is a co-recipient of the best research proposal award from SK Telecom, Korea. He is currently serving as Technical Editor of several prestigious Journals/Magazines like \textit{IEEE Transactions on Vehicular Technology}, \textit{Elsevier Computer and Electrical Engineering}, \textit{Springer Nature Wireless Personal Communications}, \textit{Human-centric Computing and Information Sciences}, \textit{Journal of Information Processing Systems}, and \textit{Frontiers in Communications and Networks}. He has served/serving as Guest Editor for special issues in \textit{IEEE Wireless Communications}, \textit{IEEE Communications Magazine}, \textit{IEEE Access}, \textit{Sensors}, \textit{IEEE/KICS Journal of Communications and Networks}, and \textit{Physical Communications}, and served as a Track Chair in VTC-Fall 2024 and VTC-Spring 2025. He also regularly serves as TPC Member for world-distinguished conferences like IEEE Globecom, IEEE VTC, IEEE ICC, and IEEE PIMRC.
\end{IEEEbiography}

\begin{IEEEbiography}[{\includegraphics[width=1in,height=1.25in,clip,keepaspectratio]{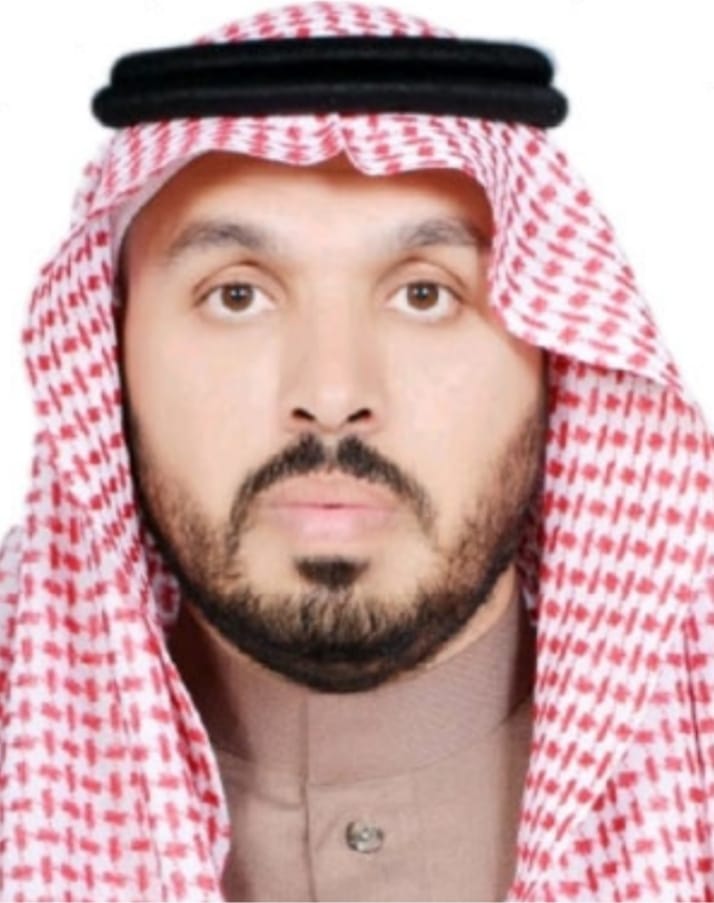}}]{Ali H. Alenezi} received the B.S. degree in
electrical engineering from King Saud University, Saudi Arabia, the M.S. degree in electrical
engineering from the KTH Royal Institute of
Technology, Sweden, and the Ph.D. degree in
electrical engineering from New Jersey Institute
of Technology, USA, in 2018. He is currently
an Associate Professor with the Electrical Engineering Department, Northern Border University,
Saudi Arabia. His research interests include acoustic communication, wireless communications, and 4G and 5G networks
using UAVs.
\end{IEEEbiography}

\begin{IEEEbiography}[{\includegraphics[width=1in,height=1.25in,clip,keepaspectratio]{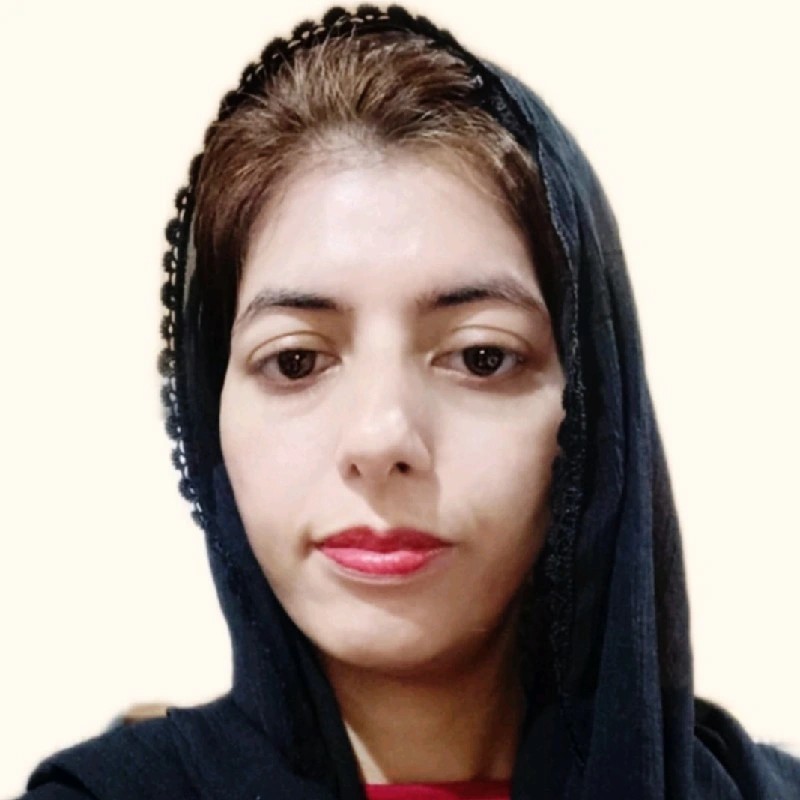}}]{Iqra Shabir} is currently enrolled in a PhD program, with research focused on Social Robot Assistant for Intelligent Healthcare in Università degli Studi di Genova, Italy. She has completed MS in Software Engineering from KFUEIT, Pakistan,  Her main research interests include Multi-Agent Reinforcement Learning (MARL), Decentralized Decision-Making, and Human-Robot Interaction in Healthcare Systems. 
\end{IEEEbiography}

\begin{IEEEbiography}[{\includegraphics[width=1in,height=1.25in,clip,keepaspectratio]{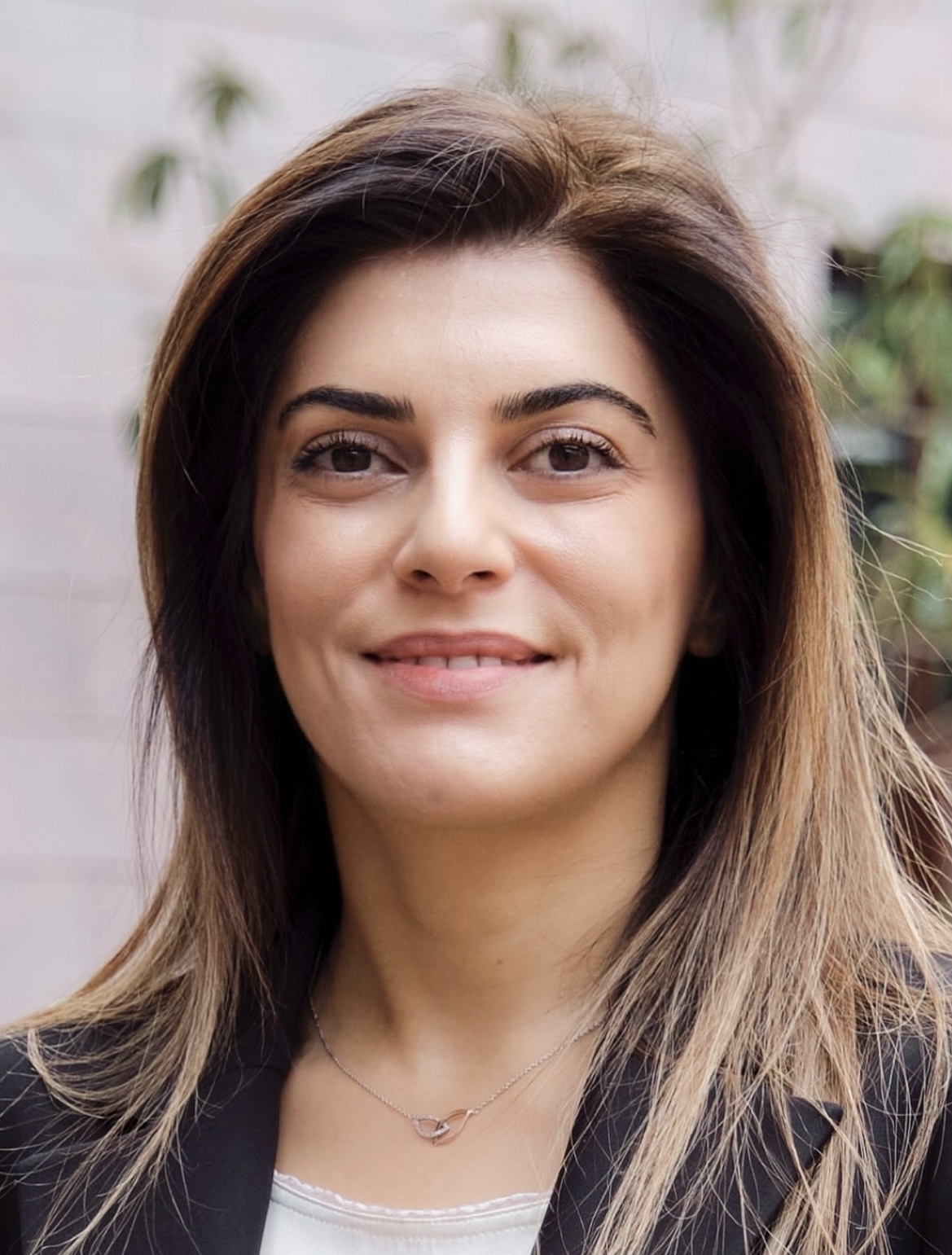}}]{Sinem Coleri} is a Professor in the Department of Electrical and Electronics Engineering at Koc University. She is also the founding director of Wireless Networks Laboratory (WNL) and director of Ford Otosan Automotive Technologies Laboratory. Sinem Coleri received the BS degree in electrical and electronics engineering from Bilkent University in 2000, the M.S. and Ph.D. degrees in electrical engineering and computer sciences from University of California Berkeley in 2002 and 2005. She worked as a research scientist in Wireless Sensor Networks Berkeley Lab under sponsorship of Pirelli and Telecom Italia from 2006 to 2009. Since September 2009, she has been a faculty member in the department of Electrical and Electronics Engineering at Koc University. Her research interests are in 6G wireless communications and networking, machine learning for wireless networks, machine-to-machine communications, wireless networked control systems and vehicular networks. She has received numerous awards and recognitions, including TUBITAK (The Scientific and Technological Research Council of Turkey) Science Award in 2024; N2Women: Stars in Computer Networking and Communications in 2022; TUBITAK Incentive Award and IEEE Vehicular Technology Society Neal Shepherd Memorial Best Propagation Paper Award in 2020; Outstanding Achievement Award by Higher Education Council in 2018; and Turkish Academy of Sciences Distinguished Young Scientist (TUBA-GEBIP) Award in 2015.  Dr. Coleri currently holds the position of Editor-in-Chief at the IEEE Open Journal of the Communications Society.  Dr. Coleri is an IEEE Fellow, AAIA Fellow and IEEE ComSoc Distinguished Lecturer.
\end{IEEEbiography}
\begin{IEEEbiography}[{\includegraphics[width=1in,height=1.25in,clip,keepaspectratio]{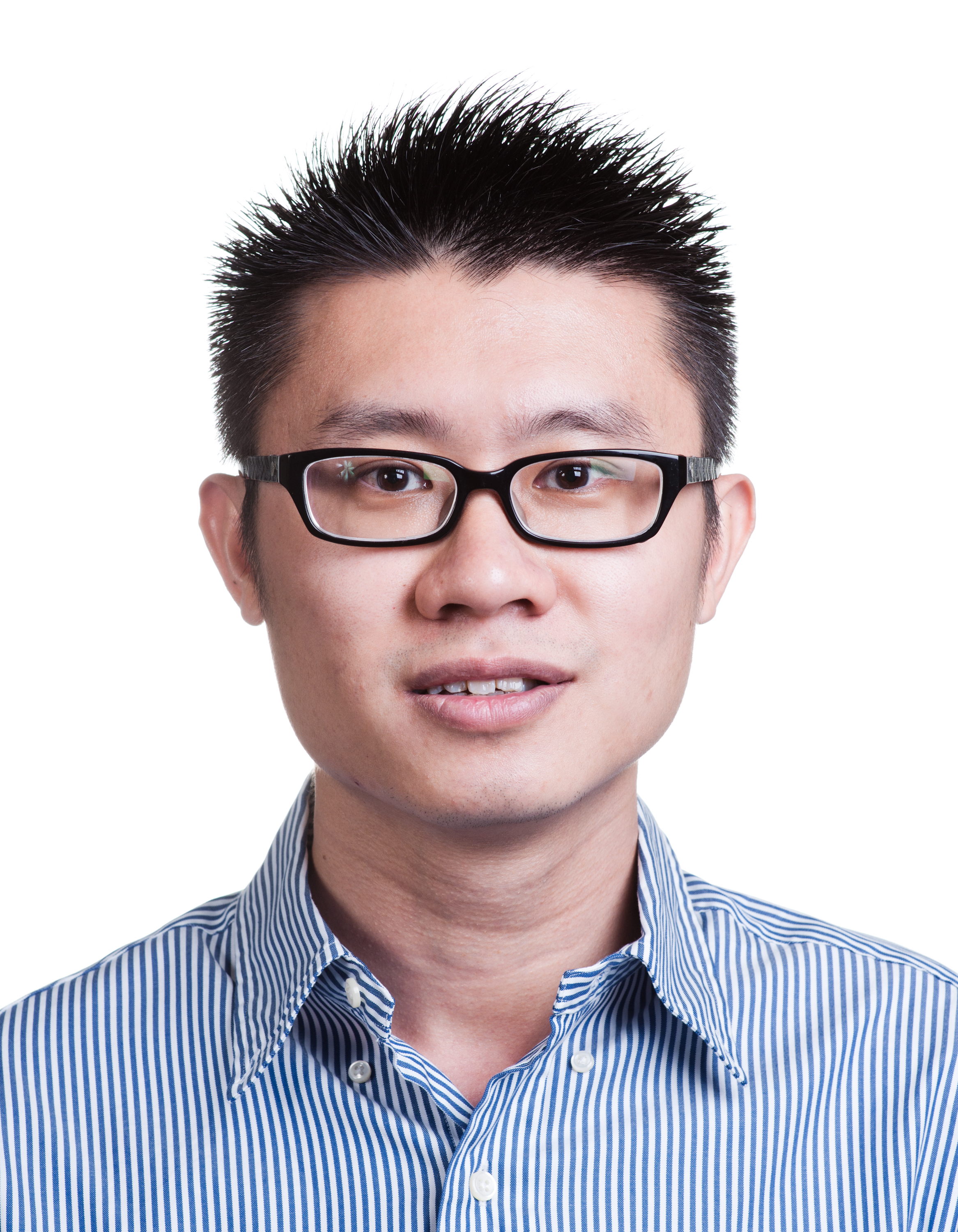}}]{Chau Yuen} (Fellow, IEEE) received the B.Eng. and Ph.D. degrees from Nanyang Technological University, Singapore, in 2000 and 2004, respectively.
He was a Postdoctoral Fellow with Lucent Technologies Bell Labs, Murray Hill, NJ, USA, in 2005, and a Visiting Assistant Professor with The Hong Kong Polytechnic University, Hong Kong in 2008. From 2006 to 2010, he was with the Institute for Infocomm Research, Singapore. From 2010 to 2023, he was with the Engineering Product Development Pillar, the Singapore University of Technology and Design, Singapore. Since 2023, he has been with the School of Electrical and Electronic Engineering, Nanyang Technological University. He has three U.S. patents and published more than 500 research papers in international journals or conferences.
Dr. Yuen received the IEEE ICC Best Paper Award in 2023, the IEEE Communications Society Fred W. Ellersick Prize in 2023, the IEEE Marconi Prize Paper Award in Wireless Communications in 2021, and the EURASIP Best Paper Award for Journal on Wireless Communications and Networking in 2021. He was a recipient of the Lee Kuan Yew Gold Medal, the Institution of Electrical Engineers Book Prize, the Institute of Engineering of Singapore Gold Medal, the Merck Sharp and Dohme Gold Medal, and twice a recipient of the Hewlett Packard Prize. He received the IEEE Asia–Pacific Outstanding Young Researcher Award in 2012 and the IEEE VTS Singapore Chapter Outstanding Service Award in 2019. He is a Distinguished Lecturer of the IEEE Vehicular Technology Society, the Top 2\% Scientists by Stanford University, and a Highly Cited Researcher by Clarivate Web of Science.
\end{IEEEbiography}

\end{document}